\documentclass{article} 
\usepackage[final]{colm2026_conference}

\usepackage[T1]{fontenc}

\usepackage{silence}
\usepackage{microtype}
\usepackage{hyperref}
\usepackage{url}

\usepackage{amsmath}
\usepackage{amssymb}

\usepackage{booktabs}
\usepackage{multirow}
\usepackage{makecell}
\usepackage{array}
\usepackage{colortbl}

\usepackage{graphicx}
\usepackage{xcolor}
\usepackage{wrapfig}
\usepackage{subcaption}
\usepackage{enumitem}
\usepackage{placeins}        

\usepackage[breakable,skins]{tcolorbox}
\usepackage{pifont}

\newcolumntype{L}[1]{>{\raggedright\arraybackslash}p{#1}}
\newcolumntype{C}[1]{>{\centering\arraybackslash}p{#1}}

\definecolor{localyellow}{RGB}{255,249,196}
\definecolor{sectiongray}{RGB}{240,240,240}
\definecolor{ocyellow}{RGB}{255,253,220}
\definecolor{paperblue}{RGB}{219,234,254}
\definecolor{lbgreen}{RGB}{220,245,220}
\definecolor{snGreen}{RGB}{131,198,97}

\newcommand{\na}{---}

\newcommand{\synthdocbench}[1]{\textsc{SynthDocBench}}

\newtcolorbox{promptbox}[1][]{
  breakable, enhanced,
  title={#1},
  fonttitle=\small\bfseries\sffamily,
  coltitle=white,
  colbacktitle=gray!60,
  colback=gray!5,
  colframe=gray!40,
  boxrule=0.5pt,
  arc=3pt,
  left=8pt, right=8pt, top=3pt, bottom=5pt,
}

\newtcolorbox{titlebox}{
  enhanced, breakable,
  colback=white, colframe=snGreen,
  boxrule=1.3pt, arc=5mm,
  left=14pt, right=14pt, top=12pt, bottom=12pt,
}

\usepackage{lineno}

\definecolor{darkblue}{rgb}{0, 0, 0.5}

\hypersetup{colorlinks=true, citecolor=darkblue, linkcolor=darkblue, urlcolor=darkblue}

\definecolor{sectiongray}{RGB}{230,230,230}   
\definecolor{localyellow}{RGB}{255,248,196}   
\definecolor{natext}{RGB}{150,150,150}        

\title{SynthDocBench: Controlled Benchmark for Long-Context Visual Document Understanding}

\hypersetup{
  pdftitle={SynthDocBench: Controlled Benchmark for Long-Context Visual Document Understanding},
  pdfauthor={Abhigya Verma, Khyati Mahajan, Amit Kumar Saha, Shruthan Radhakrishna, Sagar Davasam, Vikas Yadav, Sai Rajeswar}
}

\begin{document}

\ifcolmsubmission
\linenumbers
\fi

\lhead{Published as a conference paper at COLM 2026}
\setcounter{footnote}{0}

\begin{titlebox}
\raisebox{-.65ex}{\includegraphics[width=0.78cm]{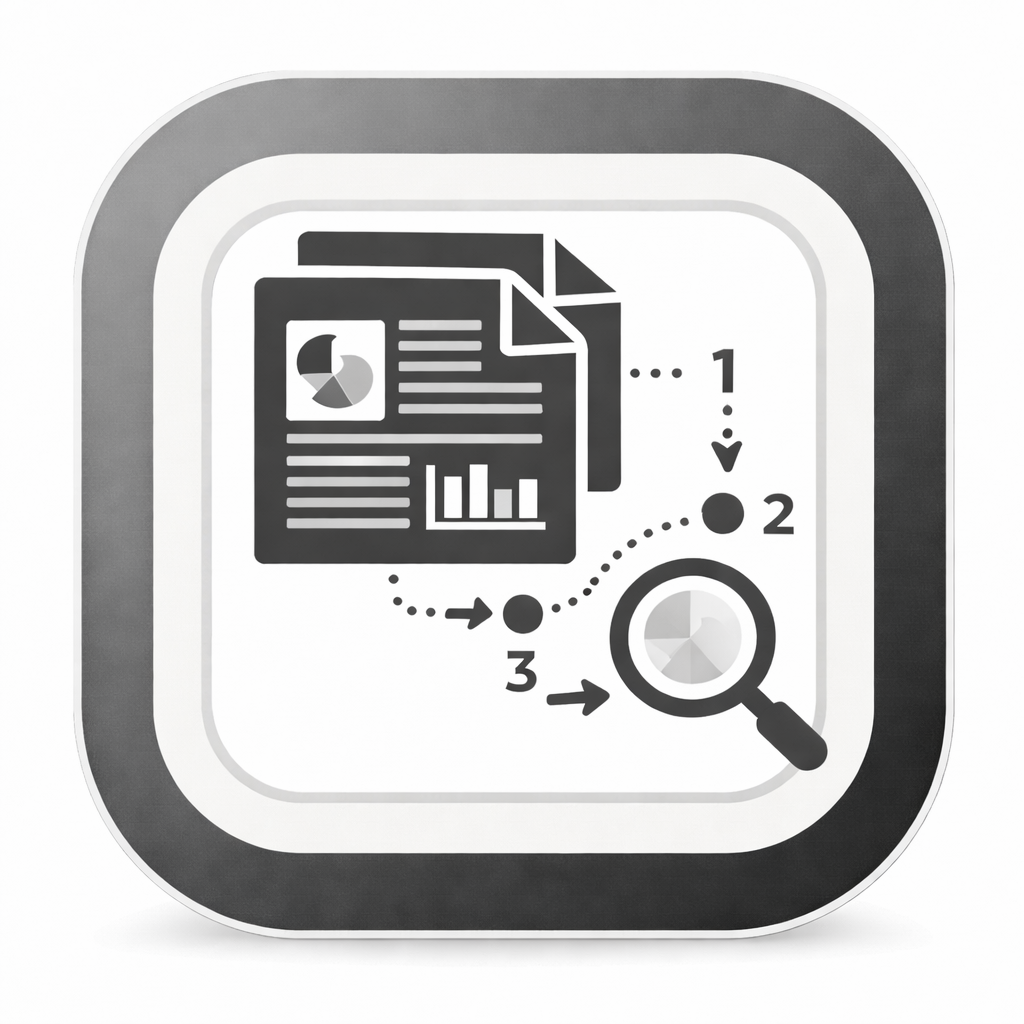}}\,%
{\Large\bf\textls[-5]{SynthDocBench: Controlled Benchmark for Long-Context Visual Document Understanding}}

\vspace{8pt}
\textbf{Abhigya Verma}\textsuperscript{\rm 1*}\,,
\textbf{Khyati Mahajan}\textsuperscript{\rm 1*}\,,
\textbf{Amit Kumar Saha}\textsuperscript{\rm 1*}\,,
\textbf{Shruthan Radhakrishna}\textsuperscript{\rm 1}\,,
\textbf{Sagar Davasam}\textsuperscript{\rm 1}\,,
\textbf{Vikas Yadav}\textsuperscript{\rm 1}\,,
\textbf{Sai Rajeswar}\textsuperscript{\rm 1, 2, 3}

\vspace{4pt}
\textsuperscript{\rm 1}ServiceNow AI\quad
\textsuperscript{\rm 2}Mila\quad
\textsuperscript{\rm 3}Universit\'e de Montr\'eal

\vspace{10pt}

Vision language models (VLMs) have achieved strong performance on visual document understanding benchmarks such as DocVQA, ChartQA, and MMLongBench-Doc. However, real-world documents combine multiple factors such as length, layout complexity, modality, and question difficulty, which makes it difficult to attribute model failures to specific causes.
We introduce \textsc{SynthDocBench}, a fully synthetic benchmark for long-context visual document understanding that systematically controls factors including document length, layout structure, modality composition, and question type. The benchmark is constructed using a combinatorial design, each factor is varied independently across generated documents, enabling controlled analysis of model behavior. Documents are generated end to end using an LLM pipeline across six layout archetypes, with a 40 percent random override to prevent models from exploiting spurious correlations. Additionally, SynthDocBench spans long-context documents with substantially greater length and structural diversity than existing benchmarks.
Evaluating seven frontier VLMs, we uncover three failure modes that existing benchmarks cannot surface: sharp degradation with document length, a systematic positional sensitivity in which the middle third of a document is hardest for five of six models and five of six models show a negative Early$\to$Late trend (steepest decline: 8.3 percentage points), and breakdown of chart comprehension in long-document settings. These results suggest that current models may be overfitting to benchmark artifacts rather than achieving robust long-context visual document understanding.

\vspace{10pt}
\noindent
\begin{minipage}[b]{0.72\textwidth}
\footnotesize
\textbf{Correspondence:} \href{mailto:abhigya.verma@servicenow.com}{abhigya.verma@servicenow.com}\\[2pt]
\textbf{Code:} \url{https://github.com/ServiceNow/SynthDocBench}\quad
\textbf{Dataset:} \url{https://huggingface.co/datasets/ServiceNow-AI/SynthDocBench}\\[4pt]
{\scriptsize \textsuperscript{\rm *}Equal contribution.}
\end{minipage}%
\hfill
\begin{minipage}[b]{0.22\textwidth}
\raggedleft
\includegraphics[width=0.9\textwidth]{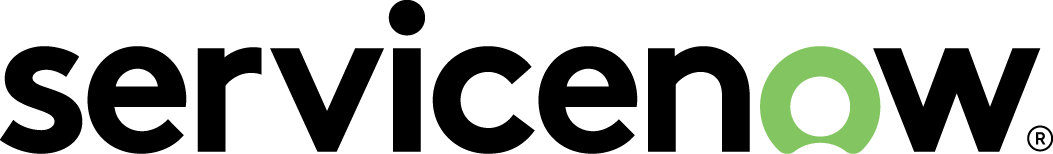}
\end{minipage}
\end{titlebox}

\section{Introduction}

Understanding long, visually rich documents is a defining challenge for vision language models (VLMs). Real-world documents interleave text, tables, charts, and complex layouts across dozens or hundreds of pages, demanding both long-range retrieval and cross-modal reasoning. Benchmarks such as DocVQA~\citep{mathew2021docvqa}, ChartQA~\citep{masry2022chartqa}, and MMLongBench-Doc~\citep{ma2024mmlongbenchdoc} have driven substantial progress, yet this progress conceals a fundamental diagnostic blind spot: \emph{when a model fails on a real document, it is difficult to know why}.

On single-page tasks, evaluation is approaching saturation. Frontier models exceed 95\% on DocVQA~\citep{bai2025qwen3,wang2025internvl3} and 89\% on ChartQA~\citep{anthropic2024claude45sonnet,bai2025qwen3} --- though harder benchmarks such as ChartQAPro~\citep{masry2025chartqapro}, VisuLogic~\citep{zhang2025visulogic}, and ChartMuseum~\citep{xu2025chartmuseum} show that chart understanding is far from solved, with degradations exceeding 30 percentage points. Yet all evaluate charts in isolation, abstracted from the multi-page contexts in which they naturally occur. Long-context benchmarks like  MMLongBench-Doc \citep{ma2024mmlongbenchdoc} address the document-length dimension, targeting long PDF comprehension with rich visual content, and the strongest model achieves only 57\%. LongDocURL~\citep{deng2025longdocurl} and M-LongDoc~\citep{chia2024mlongdoc} extend to documents spanning 100s of pages, but prioritize breadth of coverage over controlled diagnosis: neither constructs questions requiring joint reasoning over charts and textually distributed evidence across distant pages. Furthermore, these benchmarks draw on real documents, confounding potential sources of difficulty (answer depth, presentation modality, layout density, cross-page evidence integration) which co-vary and cannot be disentangled.

Synthetic benchmarks have a long history of enabling precisely this kind of decomposition. The bAbI tasks~\citep{weston2015babi} and SCAN~\citep{lake2018scan} used programmatic generation to isolate reasoning primitives in NLP; CLEVR~\citep{johnson2017clevr} and RAVEN~\citep{zhang2019raven} did the same for visual relational and analogical reasoning; and PuzzleVQA~\citep{chia2024puzzlevqa} recently applied synthetic abstract patterns to diagnose multimodal reasoning bottlenecks in VLMs. The common principle is that synthetic control trades ecological validity for interpretability, enabling attribution of failures to specific causes rather than an opaque bundle of confounds. No comparable instrument exists for long-context visual document understanding, an important setting where confounds are arguably most severe.

We address this gap with \textbf{\synthdocbench{}}, a fully synthetic, controlled benchmark that enables the first systematic decomposition of VLM failure modes in long-context document understanding by varying document length, page depth, modality composition, and question type as independent axes. We contribute:

\begin{enumerate} [nolistsep, noitemsep, leftmargin=12pt]
    \item \synthdocbench{}, a controlled synthetic benchmark with independently variable length, depth, modality, and question-type axes, released as three task-specific subsets, which probes 24 distinct D3.js chart types (including visually ambiguous forms such as dumbbell, lollipop, slope, and sparkline grids) requiring exact numerical reads that cannot be inferred from surrounding text; \texttt{cross\_modal}, which places supporting charts and corroborating text in separate, non-adjacent sections to test cross-modal grounding over long-range dependencies; and \texttt{complex}, which requires combining 2 to 4 evidence units from text and charts across difficulty levels L1 to L5, ranging from direct value lookup to cross-section synthesis. 
    \item A fully automated LLM-based generation pipeline with a \emph{dual-layer document design}: every chart is generated simultaneously as a rendered D3.js visualisation and as a hidden structured metadata block used only for ground-truth derivation, making answers deterministic by construction across 24 chart types and 6 layout archetypes.
    \item A systematic empirical analysis of seven frontier VLMs, validated with a cross-judge robustness check (GPT-5 and Gemini-as-judge agree within 3.5 ACC points, $r \geq 0.94$, across all question types), revealing three concurrent, previously unobservable failure modes: sharp performance degradation with increasing evidence complexity and reasoning depth (L1$\to$L5), systematic positional sensitivity in which the middle section of a document is hardest for five of six models and five of six models exhibit a negative Early$\to$Late trend (steepest decline: 8.3 pp), and collapse of precise chart-reading accuracy in long-document contexts, even for models that perform well on existing benchmarks.
\end{enumerate}

\section{Related Work}
\label{sec:related}

\begin{figure}[t!]
\begin{minipage}{\textwidth}
  \centering
  \includegraphics[width=0.55\textwidth]{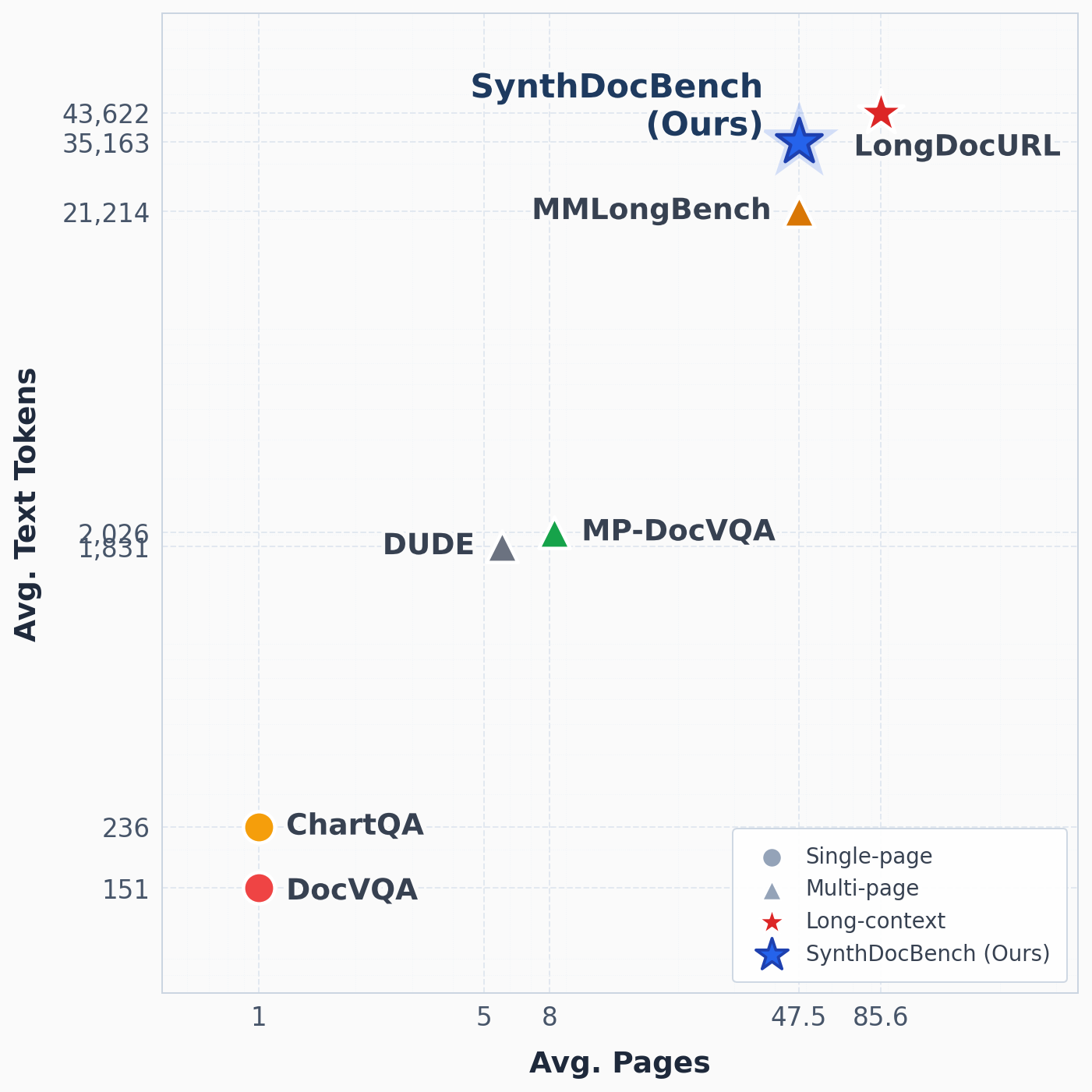}
\end{minipage}

\vspace{14pt}

\begin{minipage}{\textwidth}
  \centering
  \footnotesize
  \resizebox{\textwidth}{!}{%
  \begin{tabular}{@{}lcccc@{}}
  \toprule
  \textbf{Benchmark} & \textbf{Scope} & \textbf{Avg.\ Context} & \textbf{\#Docs/Charts} & \textbf{\#Questions} \\
  \midrule
  \rowcolor{gray!15} \multicolumn{5}{c}{{Chart Benchmarks}} \\
  \midrule
  ChartQA \citep{masry2022chartqa}               & Isolated charts   & 1 chart       & 21,953    & 32,701 \\
  ChartQAPro \citep{masry2025chartqapro}         & Isolated charts   & 1 chart       & 1,341     & 1,948  \\
  ChartGalaxy \citep{li2025chartgalaxy}          & Isolated charts   & 1 chart       & 1,763,189 & ---    \\
  VisuLogic \citep{zhang2025visulogic}           & Isolated images   & 1 image       & 1,000     & 1,000  \\
  ChartMuseum \citep{xu2025chartmuseum}          & Isolated charts   & 1 chart       & 928       & 1,162  \\
  MultiChartQA \citep{zhu2025multichartqa}       & Multi-chart       & 2--3 charts   & 655       & 944    \\
  \midrule
  \rowcolor{gray!15} \multicolumn{5}{c}{{Document VQA Benchmarks}} \\
  \midrule
  DocVQA \citep{mathew2021docvqa}                & Single-page docs  & 1 page        & 12,767    & 50,000 \\
  MP-DocVQA \citep{tito2023hierarchical}         & Multi-page docs   & $\leq$20 pages & ---      & 46,000 \\
  SlideVQA \citep{tanaka2023slidevqa}            & Multi-page docs   & $\sim$20 pages & 2,619    & 14,500 \\
  MMLongBench-Doc \citep{ma2024mmlongbenchdoc}   & Multi-page docs   & 47.5 pages    & 135       & 1,082  \\
  LongDocURL \citep{deng2025longdocurl}          & Multi-page docs   & $\sim$83 pages & 396      & 2,325  \\
  M-LongDoc \citep{chia2024mlongdoc}             & Multi-page docs   & $>$200 pages  & ---       & 851    \\
  MMLongBench \citep{wang2025mmlongbench}        & Multi-page docs   & 8K--128K tokens & 13,331  & ---    \\
  \midrule
  \includegraphics[width=0.8em]{images/icon2.png} \textbf{\textsc{SynthDocBench}} & \textbf{Multi-page docs} & \textbf{Avg.\ 51.1 pages} & \textbf{200 / 3,340} & \textbf{1,788} \\
  \bottomrule
  \end{tabular}
  }
\end{minipage}
\vspace{-6pt}
\caption{\textbf{Landscape of benchmarks} 
Top: benchmark comparison by average document length (pages) and average textual context (tokens). SynthDocBench occupies a unique region with both long multi-page context and high textual density.
Bottom: comparison with existing chart and document VQA benchmarks. 
Prior benchmarks typically isolate either charts or long documents, whereas \textbf{\textsc{SynthDocBench}} is designed to study their intersection under long contexts.}
\label{fig:related_benchmarks}
\end{figure} 

\paragraph{Charts and Visual Reasoning.}
Chart understanding has been evaluated primarily in isolation from document context. ChartQA~\citep{masry2022chartqa} established the standard evaluation setup for chart-oriented VQA, now approaching saturation. ChartQAPro~\citep{masry2025chartqapro} addresses this through harder, more diverse real-world charts, exposing over 30 percentage points of degradation for models that saturate ChartQA, though it retains the isolated-image setting. ChartGalaxy~\citep{li2025chartgalaxy} demonstrates that programmatic generation can yield million-scale synthetic diversity spanning 75 chart types and 330 stylistic variations, providing a methodological precedent for \synthdocbench{}'s synthetic approach. VisuLogic~\citep{zhang2025visulogic} and ChartMuseum~\citep{xu2025chartmuseum} probe fine-grained logical and perceptual reasoning over isolated charts, while MultiChartQA~\citep{zhu2025multichartqa} extends this to simultaneous multi-chart reasoning. Collectively, these benchmarks establish that chart understanding is far from solved, yet none evaluate charts within the document contexts where they naturally occur and where interpretation requires engagement with surrounding text, tables, and figures.

\paragraph{Long-Context Multimodal Benchmarks.}
Early long-context evaluation focused on needle-in-a-haystack retrieval~\citep{wang2024mmniah}, which provides insufficient difficulty ceilings for frontier models and correlates poorly with downstream reasoning performance~\citep{wang2025mmlongbench}. Document-centric benchmarks have progressed from single-page evaluation via DocVQA~\citep{mathew2021docvqa} through multi-page extensions including MP-DocVQA~\citep{tito2023hierarchical} and SlideVQA~\citep{tanaka2023slidevqa}, though both are constrained to roughly twenty pages and principally assess span extraction. MMLongBench-Doc~\citep{ma2024mmlongbenchdoc} was the first benchmark targeting long PDF comprehension with interleaved visual content, and MMLongBench~\citep{wang2025mmlongbench} extended this framework to five task categories spanning up to 128K tokens. LongDocURL~\citep{deng2025longdocurl} and M-LongDoc~\citep{chia2024mlongdoc} advance the state of the art through cross-element localisation and open-ended responses, but both prioritize breadth of coverage over diagnostic decomposition. 
\synthdocbench{} is distinguished by its explicitly diagnostic design: document length, page depth, modality composition, and question type are varied as independent axes, with charts and figures foregrounded as primary reasoning targets whose interpretation requires engagement with arbitrarily distant contextual evidence.

\paragraph{Vision-Language Models for Document Understanding.}
Recent vision-language models have made single-page document QA increasingly saturated: frontier systems such as Qwen3-VL~\citep{bai2025qwen3} and InternVL3.5~\citep{wang2025internvl3} now approach ceiling performance on DocVQA, so these benchmarks provide limited diagnostic separation among leading models. In contrast, long-context visual document reasoning remains substantially harder: the best reported results on MMLongBench-Doc remain far below single-page performance.\footnote{\url{https://huggingface.co/spaces/OpenIXCLab/mmlongbench-doc}, accessed March 2026.} This gap suggests that failures are driven by properties that emerge in multi-page settings e.g., long-range dependency tracking, retrieval over dispersed evidence, yet current evaluations do not isolate which document factors are most responsible. Figure~\ref{fig:related_benchmarks} situates document VQA benchmarks by average document length (pages) and average token count. Single-page benchmarks (DocVQA, ChartQA) lie in the bottom-left, whereas multi-page benchmarks shift toward higher length and token regimes. \synthdocbench{} lies near the frontier along both axes, comparable to LongDocURL in page count and substantially denser in tokens than MMLongBench, while using controlled synthetic documents and requiring retrieval and reasoning over multi-modal evidence across multiple pages.

\section{The \synthdocbench{} Benchmark}
\label{sec:methodology}

The benchmark is generated via three coupled stages (Figures~\ref{fig:report_pipeline} and \ref{fig:qa_pipeline}): (i)~\emph{document generation}, mapping a topic seed to a styled visual report with an aligned structured manifest; (ii)~\emph{QA generation}, converting the manifest into difficulty-controlled QA pairs; and (iii)~\emph{vision-only evaluation}, measuring model performance on rendered page images with no access to the underlying metadata. Reference answers are derived deterministically from the same structured artifacts used to generate the documents, eliminating the annotation bottleneck of real-document benchmarks~\citep{weston2015babi,johnson2017clevr}.

\subsection{Synthetic Visual Document Generation}
\label{sec:pipeline:report}

\begin{figure}[t]
  \centering
  \vspace{4pt}
  \includegraphics[width=\textwidth]{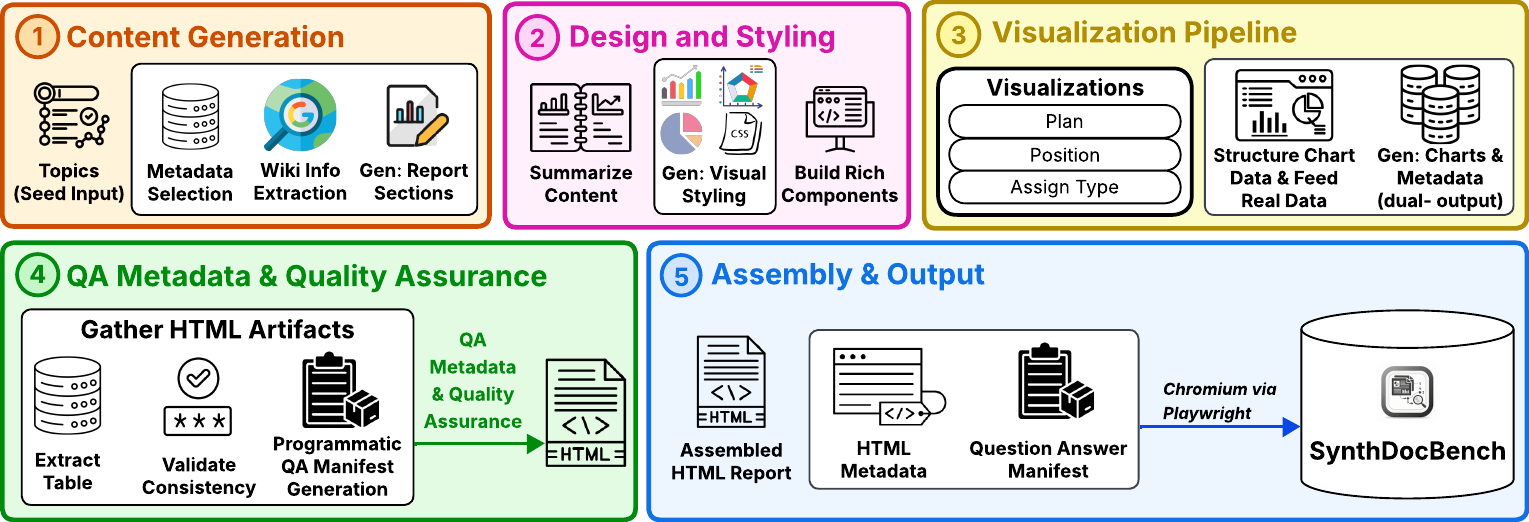}
  \vspace{4pt}
  \caption{
    \textbf{Synthetic visual document generation pipeline.}
   From a topic seed, the pipeline generates grounded report content, applies document-level visual styling, synthesizes visualizations, performs metadata and QA validation, and assembles the final HTML/PDF reports with a machine-readable QA manifest.
  }
  \label{fig:report_pipeline}
\end{figure}

The document-generation pipeline maps a topic seed $\tau$ to a styled visual report $\mathcal{D}$ and a document-level QA manifest $\mathcal{M}$, factorizing synthesis into content generation, visual grammar, visualization synthesis, and assembly (Figure~\ref{fig:report_pipeline}).

\textbf{Topic-grounded content generation:} Given $\tau$, the pipeline constructs a semantic backbone from retrieved topic-relevant evidence and reorganizes it into a structured intermediate representation that exposes section boundaries, data-bearing spans, and salient content units. Downstream stages operate over this representation rather than free-form text, enabling precise evidence tracing. \\
\textbf{Design and visual grammar:} A layout archetype $a \in \mathcal{A}$ is sampled with probability 0.6 from a topic-conditioned distribution and with probability 0.4 uniformly at random (see Appendix~\ref{app:archetypes}), preventing trivial correlations between subject matter and layout. The archetype governs page grammar, chart placement strategy, and auxiliary component types (metric cards, timelines, pull quotes), yielding a realistic and diverse report distribution. \\
\textbf{Grounded visualization synthesis:} Each visualization is generated in two aligned forms: (i)~a visible D3.js rendering that appears in the document, and (ii)~a structured metadata object $V_k$ recording chart semantics (axes, data points, derived insights). This \emph{dual-layer} formulation ensures ground truth is available by construction without post-hoc chart parsing or human labeling (full schema in Appendix~\ref{app:viz_schema}). \\
\textbf{Validation and assembly:} Numeric values in both chart and table metadata are recomputed from structured data and corrected before finalization; all validated metadata are aggregated into $\mathcal{M}$. The report is assembled as HTML and rendered to PDF using Playwright,\footnote{\url{https://github.com/microsoft/playwright}} producing an aligned pair $(\mathcal{D}, \mathcal{M})$.

\subsection{Question-Answer Generation}
\label{sec:pipeline:qa}

\begin{figure}[t]
  \centering
  \vspace{4pt}
  \includegraphics[width=\textwidth]{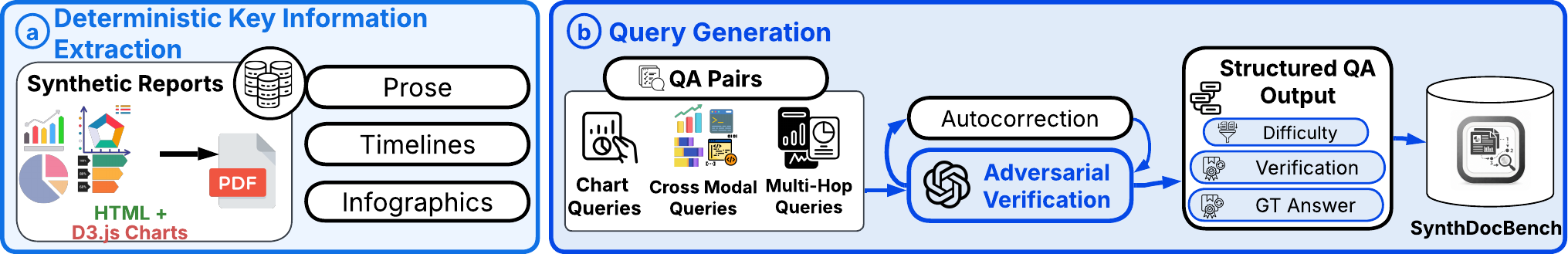}
  \vspace{4pt}
  \caption{%
    \textbf{QA generation pipeline.}
    The pipeline parses the generated report into structured evidence channels, extracts and synthesizes key information, and generates chart-reading, cross-modal, and multi-hop questions. A verification stage filters weak or malformed items before serializing the final QA output.%
  }
  \label{fig:qa_pipeline}
\end{figure}

The second stage converts a generated report into structured QA items via evidence recovery, synthesis, and generation with validation (Figure~\ref{fig:qa_pipeline}).

The pipeline parses $\mathcal{D}$ into text, table, and visualization channels, recovering chart semantics directly from embedded metadata rather than pixels. Evidence units are then synthesized into higher-order compositions supporting aggregation, comparison, and cross-source reasoning.

The benchmark produces three question families: \emph{chart-reading} (direct chart or table semantics), \emph{cross-modal} (joint reasoning over textual and visual evidence), and \emph{complex multi-hop} (composition across 2--4 evidence units). Each question carries a difficulty label $L1$--$L5$ (Table~\ref{tab:difficulty-levels}) and a structured evidence trace. A validation stage filters malformed, weakly supported, or ambiguous items; the serialized output schema is detailed in Appendix~\ref{app:qa_schema}. Three-layer quality control (numeric recomputation, automated consistency filtering, and 100-sample manual review with $>$96\% acceptance rate) is described in Appendix~\ref{app:qc}.

\subsection{Benchmark Statistics}
\label{app:doc_stats}

\begin{table}
  \centering
  \label{tab:doc_stats}
  \small
  \resizebox{0.42\textwidth}{!}{%
  \begin{tabular}{lrrrr}
    \toprule
    \textbf{Metric} & \textbf{Mean} & \textbf{Med.} & \textbf{Min} & \textbf{Max} \\
    \midrule
    Pages           & 51.1   & 49     & 24     & 91     \\
    Words           & 20,568 & 20,450 & 17,367 & 25,138 \\
    Charts          & 16.7   & 14     &  5     & 32     \\
    PDF (MB)        & 2.12   & 2.10   & 1.20   & 3.80   \\
    \bottomrule
  \end{tabular}%
  }
  \caption{Per-document statistics across all 200 reports.}
\end{table}

\begin{figure}[htbp]
  \centering
  \begin{minipage}[t]{0.62\linewidth}
    \includegraphics[width=\linewidth]{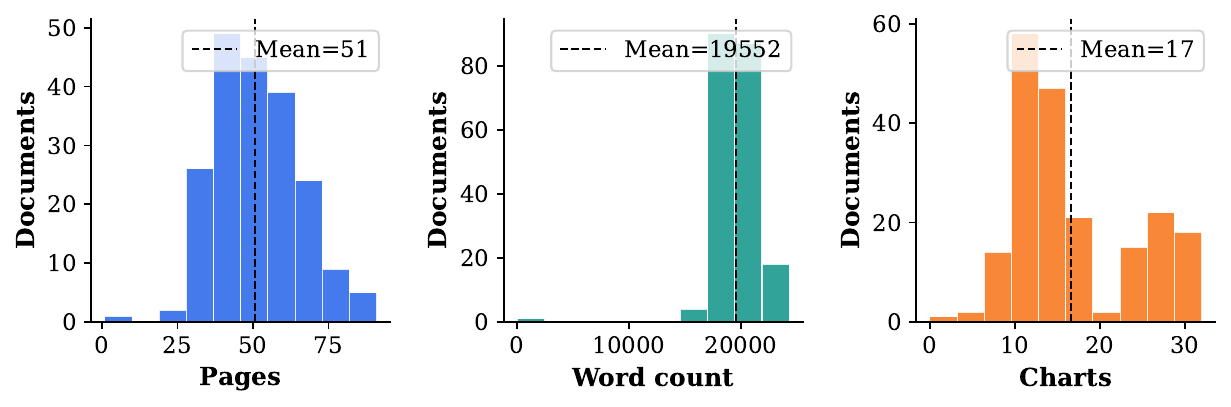}
    \vspace{4pt}
    {\small\textit{(a) Document property distributions (pages, words, charts).}}
  \end{minipage}
  \hspace{0.04\linewidth}\hfill
  \begin{minipage}[t]{0.28\linewidth}
    \includegraphics[width=\linewidth]{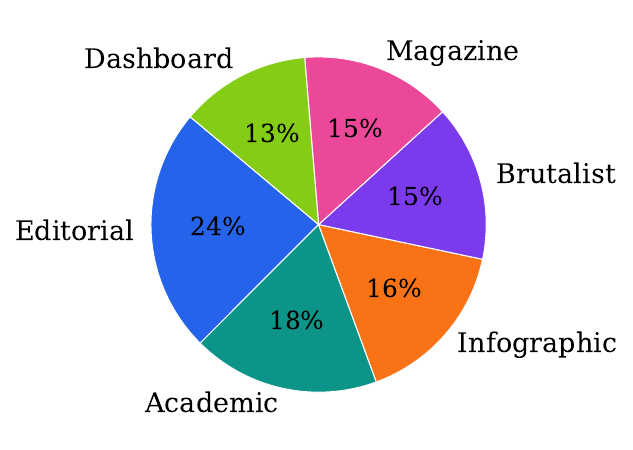}
    \vspace{4pt}
    {\small\textit{(b) Layout archetypes}}
  \end{minipage}
  \caption{Document composition statistics across 200 reports.}
  \label{fig:doc_composition}
\end{figure}

The resulting benchmark comprises \textbf{200 synthetic reports} and \textbf{1,788 questions}
distributed across three subsets: 597 \texttt{chart} reading,
597 \texttt{complex} multi-hop, and 594 \texttt{cross\_modal}.
Documents average \textbf{51.1 pages}, \textbf{16.7 charts}, and \textbf{$\approx$20,568 words},
placing \textsc{SynthDocBench} firmly in the long-document regime.
The benchmark spans \textbf{24 distinct chart types} across \textbf{6 layout archetypes},
ensuring models cannot exploit narrow visual or structural distributions.
Figure~\ref{fig:doc_composition} shows the distributions of page count, word count,
and chart count per document: all three are unimodal and tightly concentrated,
reflecting the controlled generation pipeline rather than the heavy-tailed distributions
typical of real-world corpora.
The tight ranges reflect the controlled generation pipeline: page count,
word count, and chart count are all bounded by design, enabling ablations
that hold document complexity constant while varying other axes (modality,
question type, layout).

\section{Evaluation Setup}

Candidate models operate under a strict \emph{vision-only} protocol: they receive only the rendered page-image sequence $\mathcal{I}$ and never access HTML source, embedded metadata, or $\mathcal{M}$. The end-to-end pipeline is illustrated in Figure~\ref{fig:eval_diagram} (Appendix~\ref{app:eval_pipeline}).
Each PDF is rasterized at 144\,DPI, capped at 120 pages, and concatenated into single-column 5-page vertical strips (max 7{,}900\,px, 4\,MB); we use 5-page strips as the default to satisfy the input constraints of all evaluated models simultaneously, noting that denser strips further improve performance where API limits permit. Full hyperparameters are in Table~\ref{tab:rendering_appendix}. Candidate models predict $\hat{a} = f_\theta(\mathcal{I}, q)$ at temperature~0 via a fixed system prompt requiring concise, vision-grounded 2 to 4 sentence answers (prompt in Appendix~\ref{app:prompts}). GPT-5~\citep{openai2025gpt5} scores each $(\hat{a}, a^*)$ pair at temperature~0, returning a JSON score in $[0,10]$ (rubric in Table~\ref{tab:rubric}). Parse failures receive score $-1$ and are excluded from all aggregates ($Q_{\mathrm{valid}}$). To validate judge reliability, we re-scored a subset of responses using Gemini-3.1-Pro and Claude-Sonnet-4.5 as alternative judges (Appendix~\ref{app:judge_validation}): GPT-5 and Gemini-as-judge agree to within 3.5 ACC points (Pearson $r \geq 0.94$ across all question types), confirming that model rankings are robust to judge choice (see Table~\ref{tab:judge_agreement_by_type}).

We report mean judge score and threshold accuracy:
\begin{equation}
  \text{ACC}(f_\theta) = \frac{1}{|Q_{\mathrm{valid}}|}
    \sum_{q \in Q_{\mathrm{valid}}}
    \mathbf{1}\!\left[\mathcal{J}(\hat{a}_q, a^*_q) \ge 6\right],
  \label{eq:acc}
\end{equation}
where $\tau{=}6$ (``core answer correct'') aligns with MMLongBench-Doc~\citep{ma2024mmlongbenchdoc}. Both metrics are stratified by question family and difficulty level $L1$--$L5$.

\subsection{Experiments and Results}
\label{sec:main_results}

We evaluate the following eight vision-language models on \synthdocbench{}:
Gemini-3.1-Pro~\citep{geminiteam2026gemini31},
GPT-5.4~\citep{openai2025gpt5},
GPT-4o~\citep{openai2024gpt4o},
Claude-Sonnet-4.5~\citep{anthropic2024claude45sonnet},
Qwen3.5-VL-122B~\citep{qwen2025qwen35vl},
Qwen3-VL-235B~\citep{bai2025qwen3},
InternVL3-78B~\citep{wang2025internvl3}, and
Qwen2.5-VL-7B~\citep{bai2025qwen25vl}.
All models are evaluated with GPT-5 as judge.

\begin{table*}[!htbp]
\centering
\small
\resizebox{\textwidth}{!}{%
\setlength{\tabcolsep}{3.5pt}
\begin{tabular}{@{}lcc ccccccccc@{}}
\toprule
 & \multicolumn{2}{c}{\textbf{Prior Benchmarks}}
 & \multicolumn{8}{c}{\textbf{\textsc{SynthDocBench} (Ours)}} \\
\cmidrule(lr){2-3}\cmidrule(lr){4-11}
 & \multicolumn{2}{c}{\textbf{Accuracy}}
 & \multicolumn{2}{c}{\textbf{Overall}}
 & \multicolumn{2}{c}{\textbf{Chart}}
 & \multicolumn{2}{c}{\textbf{Complex}}
 & \multicolumn{2}{c}{\textbf{Cross-Modal}} \\
\cmidrule(lr){2-3}\cmidrule(lr){4-5}\cmidrule(lr){6-7}\cmidrule(lr){8-9}\cmidrule(lr){10-11}
\textbf{Model}
  & \makecell{\textbf{DocVQA}\\\scriptsize(1\,pg)}
  & \makecell{\textbf{MMLong-}\\\textbf{Bench-Doc}\\\scriptsize(47.5\,pg)}
  & \textbf{ACC} & \textbf{Score}
  & \textbf{ACC} & \textbf{Score}
  & \textbf{ACC} & \textbf{Score}
  & \textbf{ACC} & \textbf{Score} \\
\midrule
\rowcolor{gray!15}
\multicolumn{11}{c}{\textit{Proprietary Models}} \\
\midrule
Gemini-3.1-Pro~\citep{geminiteam2026gemini31}    & 93.4          & 45.1          & \textbf{0.725} & \textbf{7.19} & \textbf{0.759} & \textbf{7.62} & \textbf{0.789} & \textbf{7.28} & \textbf{0.628} & \textbf{6.65} \\
GPT-5.4~\citep{openai2025gpt5}            & \na           & \na           & 0.423          & 4.68          & 0.425          & 4.24          & 0.457          & 5.21          & 0.387          & 4.59          \\
GPT-4o~\citep{openai2024gpt4o}            & 92.8          & 46.3          & 0.386          & 4.38          & 0.457          & 4.44          & 0.360          & 4.66          & 0.342          & 4.05          \\
Claude-Sonnet-4.5~\citep{anthropic2024claude45sonnet} & 92.0          & 40.1          & 0.314          & 3.96          & 0.353          & 4.11          & 0.337          & 4.28          & 0.250          & 3.49          \\
\midrule
\rowcolor{gray!15}
\multicolumn{11}{c}{\textit{Open-Weight Models}} \\
\midrule
Qwen3.5-VL-122B~\citep{qwen2025qwen35vl}  & \na           & \na           & 0.655          & 6.77          & 0.713          & 7.21          & 0.690          & 6.89          & 0.561          & 6.22          \\
Qwen3-VL-235B~\citep{bai2025qwen3}        & \textbf{96.5} & \textbf{57.0} & 0.586          & 6.18          & 0.642          & 6.60          & 0.611          & 6.23          & 0.503          & 5.71          \\
InternVL3-78B~\citep{wang2025internvl3}    & 95.1          & 24.3          & 0.383          & 4.39          & 0.456          & 4.75          & 0.397          & 4.73          & 0.296          & 3.69          \\
Qwen2.5-VL-7B~\citep{bai2025qwen25vl}     & 93.7          & 25.1          & 0.081          & 1.08          & 0.162          & 1.76          & 0.012          & 0.37          & 0.067          & 1.12          \\
\bottomrule
\end{tabular}%
}
\caption{%
  Performance on \textsc{SynthDocBench} (200 reports, 1,788 questions) alongside two established benchmarks.
  \textbf{DocVQA}~\citep{mathew2021docvqa} tests single-page understanding;
  \textbf{MMLongBench-Doc}~\citep{ma2024mmlongbenchdoc} tests multi-page documents (avg.\ 47.5 pages).
  \na{} = not publicly reported at time of writing.
  Bootstrap 95\% CIs (2{,}000 resamples, seed 42) are ${\leq}\pm0.023$ overall and ${\leq}\pm0.041$ per subset across all models; all pairwise ACC gaps between adjacent-ranked models exceed their combined CI half-widths.
  Per-model CI details are reported in Table~\ref{tab:bootstrap_ci}.
}
\label{tab:main_results}
\end{table*}

Table~\ref{tab:main_results} presents ACC ($\tau{=}6$) and mean judge score across all eight models, stratified by question subset. Gemini-3.1-Pro leads all models by a substantial margin, achieving an overall ACC of 0.725 and a mean judge score of 7.19. The ranking Gemini $>$ Qwen3.5-VL-122B $>$ Qwen3-VL-235B $>$ GPT-5.4 $>$ GPT-4o $\approx$ InternVL3-78B $>$ Claude-Sonnet-4.5 $\gg$ Qwen2.5-VL-7B holds consistently across all three question subsets. Qwen3.5-VL-122B ranks second overall (ACC 0.655), demonstrating that a 122B MoE open-weight model can approach Gemini-level performance on long-context document understanding. GPT-5.4 ranks fourth overall (ACC 0.423), between Qwen3-VL-235B (0.586) and GPT-4o (0.386). GPT-4o (0.386) and InternVL3-78B (0.383) are statistically indistinguishable despite their different architectures, suggesting that neither parameter scale nor training approach alone determines long-context document performance on this benchmark.

\paragraph{OCR + Text-Only Baseline.}
\begin{figure}
  \centering
  \begin{minipage}{0.48\textwidth}
    \centering
    \includegraphics[width=\textwidth]{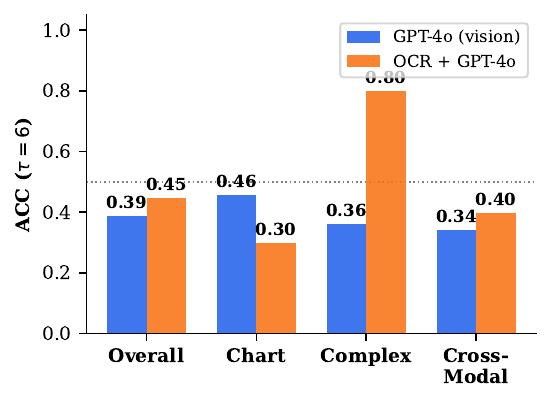}
    \caption{GPT-4o vision vs.\ OCR+GPT-4o (text-only) ACC by subset.
      Vision dominates on Chart; OCR dominates on Complex.}
    \label{fig:ocr_comparison}
  \end{minipage}
  \hfill
  \begin{minipage}{0.48\textwidth}
    \centering
    \includegraphics[width=\textwidth]{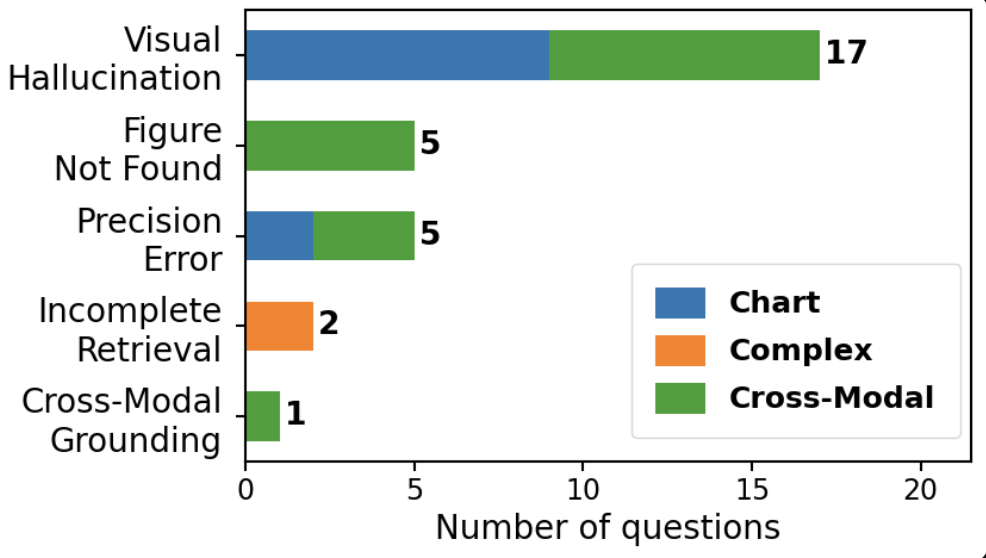}
    \caption{%
      Hard failures: 109 questions where all six models score $\leq 3$, broken down by
      error category and question subset
      (\textbf{Ch.}~= chart-reading; \textbf{Cx.}~= complex; \textbf{XM.}~= cross-modal).
    }
    \label{fig:error_analysis}
  \end{minipage}
\end{figure}

Chart-reading is the easiest subset for most models, while cross-modal questions are consistently the hardest, confirming that integrating evidence across text and charts within long documents remains an open challenge for all current VLMs. The gap between Gemini and the next-best model (Qwen3.5-VL-122B, 0.655) is 7.0 ACC points; we discuss a potential distribution-familiarity confound in Section~\ref{sec:analysis}. Model rankings on \synthdocbench{} correlate with MMLongBench-Doc (Spearman $\rho{=}0.657$, Pearson $r{=}0.683$), supporting external validity of the benchmark ordering (Appendix~\ref{app:judge_validation}).

PyMuPDF extracts page text (up to 60{,}000 chars) and GPT-4o answers without images ($n{=}1{,}525$; 263 parse failures excluded). As Figure~\ref{fig:ocr_comparison} shows, performance is highly asymmetric: complex multi-hop ACC is 0.798 for OCR vs.\ 0.360 for vision, confirming that complex evidence is largely text-recoverable; chart-reading reverses sharply (OCR 0.297 vs.\ vision 0.457), confirming that chart questions genuinely require pixel-level visual decoding. The 46 pp gap between OCR chart ACC and Gemini's (0.759) isolates visual perception as the primary bottleneck.
\section{Analysis and Discussion}
\label{sec:analysis}
\begin{table}[!htbp]
\centering
\small
\resizebox{\columnwidth}{!}{%
\begin{tabular}{@{}lcccccccccc@{}}
\toprule
 & \multicolumn{2}{c}{\textbf{L1}}
 & \multicolumn{2}{c}{\textbf{L2}}
 & \multicolumn{2}{c}{\textbf{L3}}
 & \multicolumn{2}{c}{\textbf{L4}}
 & \multicolumn{2}{c}{\textbf{L5}} \\
\cmidrule(lr){2-3}\cmidrule(lr){4-5}\cmidrule(lr){6-7}\cmidrule(lr){8-9}\cmidrule(lr){10-11}
\textbf{Model} & ACC & Sc. & ACC & Sc. & ACC & Sc. & ACC & Sc. & ACC & Sc. \\
\midrule
\rowcolor{gray!15}
\multicolumn{11}{c}{\textit{Proprietary Models}} \\
\midrule
Gemini-3.1-Pro    & \textbf{0.784} & \textbf{7.84} & \textbf{0.731} & \textbf{7.53} & \textbf{0.675} & \textbf{6.91} & \textbf{0.766} & \textbf{7.19} & \textbf{0.670} & \textbf{6.48} \\
GPT-5.4           & 0.231 & 2.37 & 0.422 & 4.66 & 0.501 & 5.29 & 0.489 & 5.28 & 0.259 & 4.02 \\
GPT-4o            & 0.271 & 2.68 & 0.455 & 4.78 & 0.451 & 4.83 & 0.400 & 4.64 & 0.173 & 3.54 \\
Claude-Sonnet-4.5 & 0.382 & 3.81 & 0.300 & 3.96 & 0.312 & 3.99 & 0.360 & 4.28 & 0.154 & 3.18 \\
\midrule
\rowcolor{gray!15}
\multicolumn{11}{c}{\textit{Open-Weight Models}} \\
\midrule
Qwen3.5-VL-122B   & 0.707 & 7.16 & 0.673 & 7.11 & 0.639 & 6.66 & 0.683 & 6.76 & 0.528 & 6.02 \\
Qwen3-VL-235B     & 0.648 & 6.48 & 0.595 & 6.55 & 0.568 & 6.01 & 0.619 & 6.25 & 0.457 & 5.36 \\
InternVL3-78B     & 0.437 & 4.42 & 0.382 & 4.35 & 0.390 & 4.48 & 0.427 & 4.63 & 0.198 & 3.62 \\
Qwen2.5-VL-7B     & 0.015 & 0.17 & 0.123 & 1.48 & 0.155 & 1.88 & 0.033 & 0.71 & 0.005 & 0.26 \\
\bottomrule
\end{tabular}%
}
\caption{%
  Model performance stratified by difficulty level $L1$--$L5$.
  \textbf{ACC} = fraction of questions with judge score $\ge 6$;
  \textbf{Sc.} = mean judge score (0--10).
  Best per column in \textbf{bold}.
  Bootstrap 95\% CIs (2{,}000 resamples, seed 42) are ${\leq}\pm0.070$ per level for all models;
  all adjacent-ranked model gaps exceed their combined CI half-widths.
  Per-level CI details are in Table~\ref{tab:bootstrap_ci_diff}.
}
\label{tab:difficulty_results}
\end{table}

We analyse model performance across three dimensions: difficulty level, question category, and error type, surfacing failure modes hidden by aggregate scores.

\paragraph{Difficulty-Stratified Results}
\label{sec:difficulty_results}
\begin{figure}[t]
  \centering
  \includegraphics[width=\textwidth]{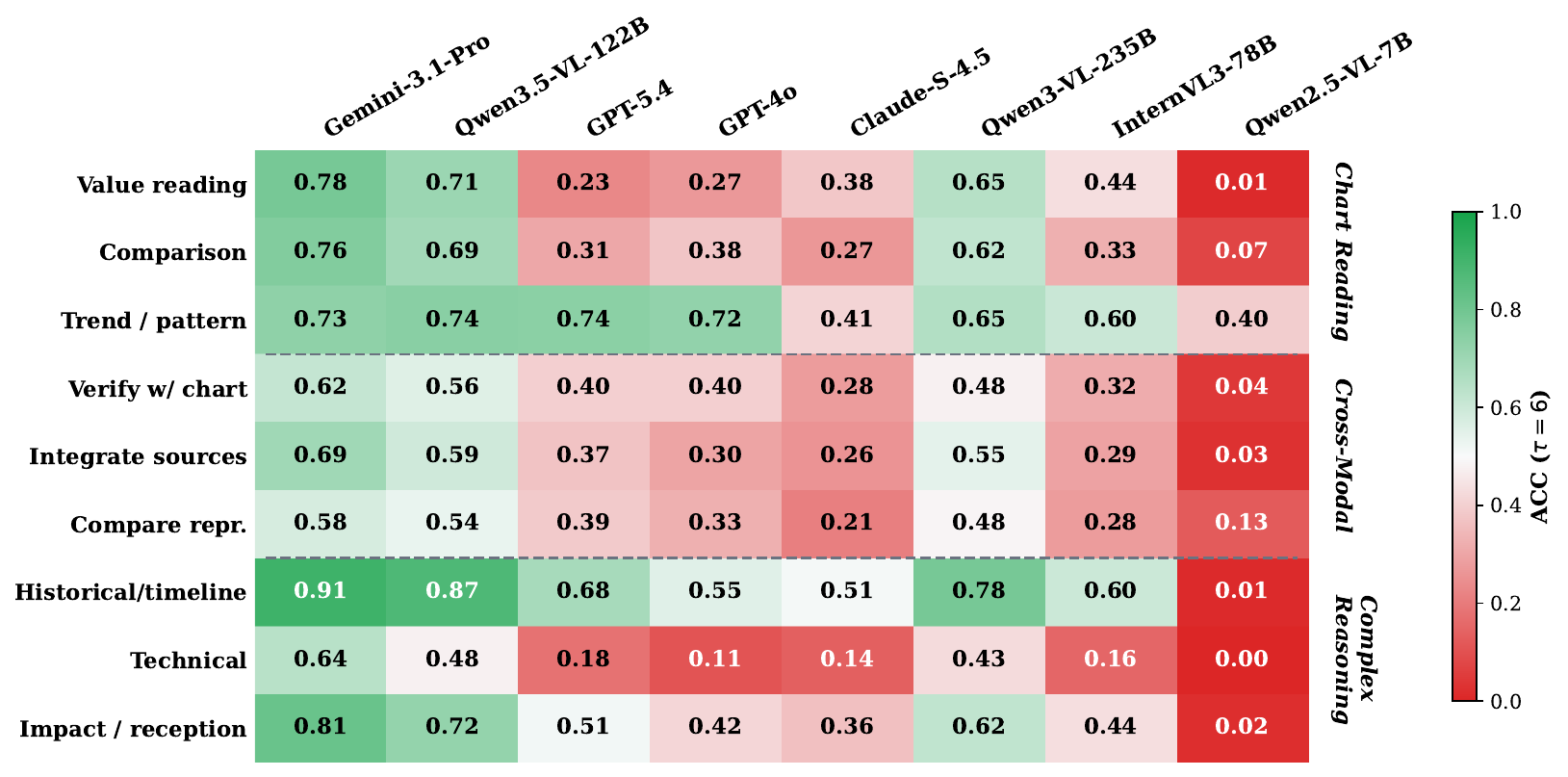}
  \caption{%
    ACC ($\tau{=}6$) by fine-grained question category.
    Rows are grouped by subset (Chart Reading, Cross-Modal, Complex Reasoning);
    columns are models ordered by overall ACC.
    The colour scale runs from red (0) to green (1).
    Full numerical values are in Table~\ref{tab:category_result_breakdown}
    (Appendix~\ref{app:category_table}).
  }
  \label{fig:category_heatmap}
\end{figure}

Table~\ref{tab:difficulty_results} shows ACC by level $L1$--$L5$. All models except Gemini-3.1-Pro degrade monotonically toward $L5$; Claude-Sonnet-4.5 drops 23 pp ($L1{\to}L5$) while Gemini stays flat (0.670--0.784). Qwen3.5-VL-122B follows a similar flat-then-drop pattern (0.707 at $L1$, 0.528 at $L5$), tracking closely with Gemini across all levels. GPT-4o's anomalous $L1$ dip (0.271) suggests precise value-extraction is harder for it than compositional reasoning.

\paragraph{Question-Category Breakdown}
\label{sec:category_results}

Figure~\ref{fig:category_heatmap} breaks down ACC by fine-grained category. Trend/pattern questions are easiest (perceptually salient direction cues), while value-reading and \emph{integrate-sources} cross-modal questions are hardest --- precise axis-label reading and multi-page evidence alignment each bottleneck distinct model families. The widest cross-model gap is in \emph{technical/quantitative} complex questions (Gemini: 0.643 vs.\ GPT-4o: 0.111, InternVL3: 0.156), isolating quantitative multi-step reasoning as the primary frontier differentiator.

\paragraph{Positional Bias: Evidence Location Within Documents}
\label{sec:positional_bias}

\begin{table}
  \centering
  \small
  \begin{tabular}{@{}lcccc@{}}
  \toprule
  \textbf{Model} & \textbf{Early} & \textbf{Middle} & \textbf{Late} & \textbf{$\Delta$} \\
  \midrule
  \rowcolor{gray!15} \multicolumn{5}{c}{\textit{Proprietary}} \\
  \midrule
  Gemini-3.1-Pro    & \textbf{0.788} & \textbf{0.717} & \textbf{0.784} & $-$0.004 \\
  GPT-5.4           & 0.440          & 0.419          & 0.468          & $+$0.028 \\
  GPT-4o            & 0.489          & 0.443          & 0.443          & $-$0.046 \\
  Claude-Sonnet-4.5 & 0.418          & 0.342          & 0.301          & $-$0.117 \\
  \midrule
  \rowcolor{gray!15} \multicolumn{5}{c}{\textit{Open-Weight}} \\
  \midrule
  Qwen3.5-VL-122B   & 0.820          & 0.635          & 0.660          & $-$0.160 \\
  Qwen3-VL-235B     & 0.674          & 0.578          & 0.693          & $+$0.019 \\
  InternVL3-78B     & 0.451          & 0.460          & 0.455          & $+$0.003 \\
  Qwen2.5-VL-7B     & 0.174          & 0.135          & 0.188          & $+$0.014 \\
  \midrule
  $n$               & 184            & 237            & 176            & --- \\
  \bottomrule
  \end{tabular}
  \caption{Chart-reading ACC ($\tau{=}6$) by position bucket.
    $\Delta$ = Late$-$Early; negative = early-content advantage.
    See Figure~\ref{fig:positional_bias} (Appendix~\ref{app:positional_bias}) for the bar chart.}
  \label{tab:positional_bias}
\end{table}

Questions are bucketed by relative chart position $p{=}k/K$ into equal thirds (Table~\ref{tab:positional_bias}; see Appendix~\ref{app:positional_bias} for the bar chart). The \textbf{middle third is hardest for 5 of 8 models}, dropping 5--18 pp below Early. Qwen3.5-VL-122B shows the steepest Early$\to$Middle drop ($-$18.5 pp) while partially recovering in Late; Claude-Sonnet-4.5 shows the steepest monotonic Early$\to$Late decline ($-$11.7 pp); Gemini-3.1-Pro shows a U-shaped pattern echoing the lost-in-the-middle effect~\citep{liu2023lostmiddlelanguagemodels}.

\paragraph{Hard Failure and Domain Error Analysis}
\label{sec:error_analysis}
Of 109 questions where all eight models score $\leq 3$ (Figure~\ref{fig:error_analysis}; Appendix~\ref{app:error_categories}), cross-modal failures dominate, confirming multi-modal evidence integration as the hardest challenge. 

Domain analysis (Figure~\ref{fig:domain_analysis}, Appendix~\ref{app:domain_analysis}) shows cross-modal ACC lags chart reading by 13-16 pp in every topic domain.

\paragraph{Visual hallucination dominates errors.}
Models return plausible-looking values absent from the ground-truth chart, concentrated on dense value-reading charts, dumbbell plots, and multi-series comparisons, localising the bottleneck to pixel-level decoding, not reasoning. Figure-not-found and precision errors account for the remainder; both point to precise quantitative visual-textual alignment as a second bottleneck that prompting alone does not close (Appendix~\ref{app:prompts}).

\paragraph{Effect of Image Presentation: Pages per Strip and Resolution}
\label{sec:ablation_rendering}

All ablations use \textbf{GPT-5} as judge on the full benchmark (Table \ref{tab:ablation_combined}; full details in Appendix~\ref{app:ablation_details}). Gemini-3.1-Pro ACC increases monotonically from 0.369 (1 page/strip) to 0.792 (10 pages), with cross-modal benefiting most (0.339$\to$0.707), confirming multi-page context is essential. 144\,DPI is optimal; higher resolution degrades performance due to JPEG compression under the 4\,MB API cap. GPT-4o and Claude-Sonnet-4.5 show model-specific optima at 2 and 5 pages respectively, indicating model-specific optimal context density.

\paragraph{Gemini-3.1-Pro dominance and potential confounds.}
The 13.9 pp gap between Gemini (0.725) and Qwen3-VL-235B (0.586) warrants caution: SynthDocBench uses web-rendered (HTML/D3.js) charts, which may be stylistically consistent with Gemini's training distribution. We cannot rule out rendering-familiarity as a partial confounder; future work should validate with alternative rendering backends to isolate long-context reasoning effects.

\begin{table}[htbp]
\centering
\small
\caption{%
  Rendering and prompting ablations on the full 200-report benchmark (judge: GPT-5).
}
\resizebox{\textwidth}{!}{%
\begin{tabular}{@{}lcccccccc@{}}
\toprule
 & \multicolumn{2}{c}{\textbf{Overall}}
 & \multicolumn{2}{c}{\textbf{Chart}}
 & \multicolumn{2}{c}{\textbf{Complex}}
 & \multicolumn{2}{c}{\textbf{Cross-Modal}} \\
\cmidrule(lr){2-3}\cmidrule(lr){4-5}\cmidrule(lr){6-7}\cmidrule(lr){8-9}
\textbf{Setting} & ACC & Score & ACC & Score & ACC & Score & ACC & Score \\
\midrule
\rowcolor{gray!15} \multicolumn{9}{c}{\textit{(a) Pages per strip (concat-num) --- Gemini-3.1-Pro, DPI fixed at 144}} \\
\midrule
1 (single page) & 0.369 & 4.04 & 0.451 & 4.65 & 0.318 & 3.48 & 0.339 & 3.99 \\
2 & 0.474 & 5.12 & 0.509 & 5.27 & 0.531 & 5.60 & 0.381 & 4.50 \\
5 (default) & 0.725 & 7.19 & 0.759 & 7.62 & 0.789 & 7.28 & 0.628 & 6.65 \\
10 & \textbf{0.792} & \textbf{7.80} & \textbf{0.838} & \textbf{8.32} & \textbf{0.831} & \textbf{7.70} & \textbf{0.707} & \textbf{7.37} \\
\midrule
\rowcolor{gray!15} \multicolumn{9}{c}{\textit{(b) Rasterization resolution (DPI) --- Gemini-3.1-Pro, concat-num fixed at 5}} \\
\midrule
72\,DPI & 0.686 & 6.97 & 0.753 & 7.50 & 0.732 & 7.14 & 0.572 & 6.25 \\
144\,DPI (default) & \textbf{0.725} & \textbf{7.19} & \textbf{0.759} & \textbf{7.62} & \textbf{0.789} & \textbf{7.28} & \textbf{0.628} & \textbf{6.65} \\
216\,DPI & 0.683 & 6.92 & 0.729 & 7.36 & 0.747 & 7.18 & 0.570 & 6.23 \\
\midrule
\rowcolor{gray!15} \multicolumn{9}{c}{\textit{(c) Prompting strategy --- Gemini-3.1-Pro}} \\
\midrule
Default & \textbf{0.725} & 7.19 & \textbf{0.759} & \textbf{7.62} & 0.789 & 7.28 & \textbf{0.628} & \textbf{6.65} \\
Chain-of-thought & 0.702 & 7.15 & 0.731 & 7.32 & 0.783 & 7.63 & 0.591 & 6.50 \\
No system prompt & 0.723 & \textbf{7.25} & 0.756 & 7.52 & \textbf{0.820} & \textbf{7.71} & 0.592 & 6.53 \\
\midrule
\rowcolor{gray!15} \multicolumn{9}{c}{\textit{(d) Prompting strategy --- GPT-4o}} \\
\midrule
Default & 0.386 & 4.38 & 0.457 & 4.44 & 0.360 & 4.66 & \textbf{0.342} & 4.05 \\
Chain-of-thought & 0.396 & 4.61 & 0.458 & 4.49 & 0.393 & 4.91 & 0.336 & \textbf{4.41} \\
No system prompt & \textbf{0.420} & \textbf{4.72} & \textbf{0.506} & \textbf{4.63} & \textbf{0.419} & \textbf{5.22} & 0.335 & 4.31 \\
\midrule
\rowcolor{gray!15} \multicolumn{9}{c}{\textit{(e) Prompting strategy --- Claude-Sonnet-4.5}} \\
\midrule
Default & 0.314 & 3.96 & 0.353 & 4.11 & 0.337 & 4.28 & 0.250 & 3.49 \\
Chain-of-thought & 0.369 & 4.44 & \textbf{0.403} & \textbf{4.36} & 0.363 & 4.49 & 0.340 & 4.47 \\
No system prompt & \textbf{0.422} & \textbf{4.90} & 0.392 & 4.31 & \textbf{0.515} & \textbf{5.73} & \textbf{0.357} & \textbf{4.64} \\
\midrule
\rowcolor{gray!15} \multicolumn{9}{c}{\textit{(f) Pages per strip --- GPT-4o (DPI fixed at 144)}} \\
\midrule
1 (single page) & 0.374 & 4.17 & 0.450 & 4.26 & 0.337 & 4.14 & 0.337 & 4.10 \\
2 & \textbf{0.396} & \textbf{4.39} & \textbf{0.470} & 4.41 & \textbf{0.381} & 4.63 & 0.337 & \textbf{4.13} \\
5 (default) & 0.386 & 4.38 & 0.457 & \textbf{4.44} & 0.360 & \textbf{4.66} & \textbf{0.342} & 4.05 \\
10 & 0.354 & 4.21 & 0.438 & 4.35 & 0.332 & 4.51 & 0.292 & 3.75 \\
\midrule
\rowcolor{gray!15} \multicolumn{9}{c}{\textit{(g) Pages per strip --- Claude-Sonnet-4.5 (DPI fixed at 144)}} \\
\midrule
1 (single page) & 0.258 & 3.15 & 0.303 & 3.51 & 0.236 & 2.78 & 0.235 & 3.15 \\
2 & 0.288 & 3.41 & \textbf{0.361} & 3.92 & 0.276 & 3.21 & 0.227 & 3.11 \\
5 (default) & \textbf{0.314} & \textbf{3.96} & 0.353 & \textbf{4.11} & \textbf{0.337} & \textbf{4.28} & \textbf{0.250} & \textbf{3.49} \\
10 & 0.312 & 3.70 & 0.378 & 4.01 & 0.314 & 4.06 & 0.244 & 3.03 \\
\bottomrule
\end{tabular}%
}
\label{tab:ablation_combined}
\end{table}
\section{Conclusion}
\label{sec:conclusion}

We introduce \synthdocbench{}, a long-context visual document understanding benchmark, measuring VLM performance for the ability to locate multiple key information facts over long-contexts from a given document to reason and answer a multi-step question. It is generated synthetically with controlled difficulty axes by independently varying document length, layout complexity, modality composition, and question type across a combinatorial design. Our results reveal three concurrent failure modes in frontier VLMs: sharp performance degradation with increasing evidence complexity and reasoning depth, systematic positional sensitivity in which the middle third of a document is the hardest section for four of six models and three of six models show a negative Early$\to$Late trend, with Claude-Sonnet-4.5 exhibiting the steepest monotonic decline ($-$11.7~pp) and Gemini-3.1-Pro showing a distinctive U-shaped recovery (§\ref{sec:positional_bias}), and collapse of precise chart-reading accuracy in long-document contexts. \synthdocbench{} reveals a clear disparity between benchmark performance and genuine long-context visual reasoning ability, serving as a diagnostic platform for future model development, providing clean, controlled signals needed to identify, understand, and ultimately improve model capabilities. Code and data will be publicly released upon acceptance.

As future work, we plan to extend \synthdocbench{} to broader multimodal reasoning settings, including richer document types (e.g., tables, forms, and mixed-layout reports), longer contexts, and more diverse reasoning tasks such as multi-hop aggregation and cross-document grounding.




\section*{Reproducibility Statement}
The authors are committed to aiding researchers in reproducing our benchmark and results. The evaluation source code is publicly available at \url{https://github.com/ServiceNow/SynthDocBench}, and the dataset is publicly available at \url{https://huggingface.co/datasets/ServiceNow-AI/SynthDocBench}. We also make every effort to disclose experiment hyperparameters wherever necessary throughout the paper, and in the appendix.

\section*{Ethics Statement}
 
\paragraph{Synthetic data and content generation.}
\textsc{SynthDocBench} is constructed entirely from programmatically generated synthetic
documents. No human subjects were involved, no personal data were collected, and no
real-world documents were reproduced. Document content is grounded in broad, publicly
available topic seeds (geopolitics, economics, environmental science, technology, and
related domains) and generated by large language models under structured constraints.
We reviewed generated content to confirm that it does not contain personally identifying
information, hate speech, or other harmful material. Because all textual content is
machine-generated and factually non-binding, it should not be treated as authoritative.

\paragraph{Model evaluation and API usage.}
Evaluation is performed via commercial APIs (Gemini, GPT-4o, Claude) and
open-weight models under their respective terms of service. No model was fine-tuned on
\textsc{SynthDocBench} data during the study; all evaluations use frozen model weights.
The GPT-5 judge is used solely to score candidate responses against deterministic
reference answers; the rubric and prompts are fully disclosed in the appendix to
allow independent replication.

\paragraph{Environmental impact.}
Large-scale model evaluation carries a non-trivial computational and energy cost. We
limited redundant evaluation runs through careful ablation design (Section~\ref{sec:analysis})
and report all configurations transparently so that future work can build on our
results rather than repeating them. Estimated compute was approximately 150 GPU-hours
for document rendering and 200 API-hours for model inference and judge scoring.

\paragraph{Benchmark integrity and Goodhart's Law.}
Releasing a controlled benchmark creates the risk that future models overfit to its
specific design choices (chart types, layout archetypes, question templates). We
mitigate this through (i) a 40\% random layout override that prevents spurious
topic--layout correlations and (ii) programmatic generation that admits straightforward
extension. We encourage the community to use \textsc{SynthDocBench} as a
\emph{diagnostic} tool and to contribute extensions that preserve the benchmark's
diagnostic validity rather than optimizing against its current instantiation.

\paragraph{Intended use.}
\textsc{SynthDocBench} is intended for research on long-context visual document
understanding. It is not intended for deployment in high-stakes decision-making
systems. Results on \textsc{SynthDocBench} characterize specific, controlled failure
modes and should not be generalized uncritically to real-world document understanding
performance.


\bibliography{colm2026_conference}
\bibliographystyle{colm2026_conference}

\appendix

\section{Evaluation Pipeline Diagram}
\label{app:eval_pipeline}

\begin{figure}[h!]
  \centering
  \includegraphics[width=0.78\textwidth]{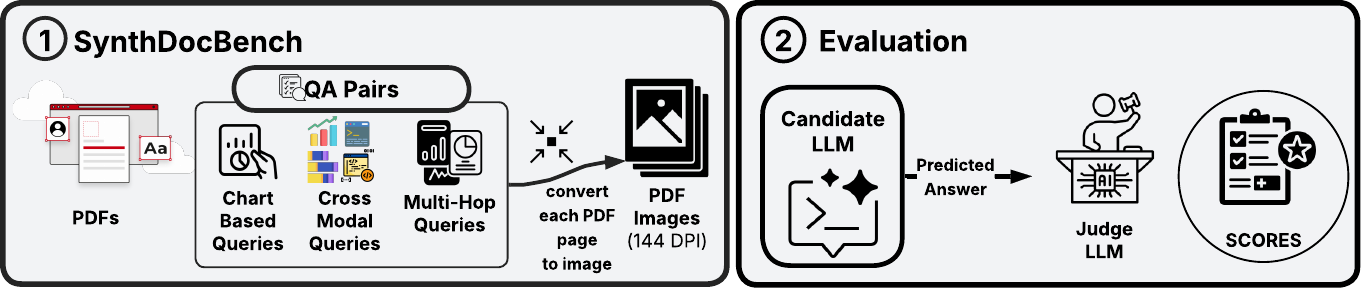}
  \caption{%
    \textbf{Evaluation pipeline.}
    Rendered PDFs are converted to page images at 144\,DPI, grouped into
    concatenated 5-page strips, and supplied directly to candidate models at
    temperature~0. Candidate answers are then scored against deterministic
    reference answers by GPT-5 acting as the judge model $\mathcal{J}$.%
  }
  \label{fig:eval_diagram}
\end{figure}

Figure~\ref{fig:eval_diagram} details the end-to-end evaluation pipeline used
to assess all candidate models on \textsc{SynthDocBench}.
Rendered PDFs are rasterized to page images at 144\,DPI, concatenated into
5-page vertical strips, and passed directly to each candidate vision-language
model at temperature~0 (vision-only; no OCR or metadata).
Each candidate response is then scored against the reference
answer by GPT-5 acting as judge $\mathcal{J}$, using the Table~\ref{tab:rubric} rubric.

\begin{table}[h!]
\centering
\small
\caption{GPT-5 judge scoring rubric. Scores of $-1$ indicate parse failures and are
excluded from aggregate statistics.}
\label{tab:rubric}
\begin{tabular}{cp{9cm}}
\toprule
\textbf{Score} & \textbf{Criterion} \\
\midrule
10   & All key facts present and fully correct \\
8--9 & Mostly correct, with only minor omissions or imprecision \\
6--7 & Core answer correct, but with some missing detail \\
4--5 & Partially correct, with significant gaps or errors \\
2--3 & Mostly incorrect or severely incomplete \\
0--1 & Incorrect, hallucinated, or contradictory to the reference \\
$-1$ & Judge parse failure (excluded) \\
\bottomrule
\end{tabular}
\end{table}

\FloatBarrier
\section{Benchmark Comparison}
\label{app:benchmark_comparison}
\begin{table*}[t]
\centering
\small
\begin{tabular}{@{}lccccc@{}}
\toprule
\textbf{Model}
  & \textbf{DocVQA}
  & \textbf{ChartQA}
  & \textbf{MathVista}
  & \textbf{MMMU}
  & \textbf{MMLongBench-Doc} \\
\midrule
\rowcolor{gray!10}
\multicolumn{6}{c}{\textit{Proprietary Models}} \\ \midrule
Gemini-3.1-Pro      & \textbf{93.4} & 88.5          & \textbf{74.8} & \textbf{81.0} & 45.1 \\
GPT-4o              & 92.8          & 85.7          & 63.8          & 70.7          & 46.3 \\
Claude-Sonnet-4.5   & 92.0          & \textbf{89.0} & 72.0          & 77.8          & 40.1 \\
\midrule
\rowcolor{gray!10}
\multicolumn{6}{c}{\textit{Open-weight Models}} \\ \midrule
Qwen3-VL-235B       & \textbf{96.5} & \textbf{91.2} & \textbf{85.8} & 71.3          & \textbf{57.0} \\
InternVL3-78B       & 95.1          & \na{}         & 79.0          & \textbf{72.2} & 24.3 \\
Qwen2.5-VL-7B       & 93.7          & 83.8          & 68.1          & 58.0          & 25.1 \\
\bottomrule
\end{tabular}
\caption{Performance of evaluated VLMs on established visual document understanding benchmarks.
  \na{} = not publicly reported at time of writing.}
\label{tab:benchmark}
\end{table*}


Table~\ref{tab:benchmark} situates \textsc{SynthDocBench} relative to the most closely
related benchmarks and reports frontier model scores on each.
Despite strong numbers on DocVQA, ChartQA, and MMLongBench-Doc, coverage is sparse and
inconsistent across models---a direct consequence of benchmarks built from heterogeneous
real corpora with no controlled variation.
Crucially, high scores on these benchmarks do not transfer: the same models that
approach saturation on DocVQA and ChartQA exhibit substantial failure rates on our
controlled subsets, demonstrating that \textsc{SynthDocBench} exposes failure modes
that existing benchmarks cannot surface.


\section{Judge Validation}
\label{app:judge_validation}

To assess the reliability of GPT-5 as judge, we re-scored the full 1{,}788-question benchmark for four candidate--judge pairs using Gemini-3.1-Pro and Claude-Sonnet-4.5 as alternative judges. The same two candidates (GPT-4o and Qwen3-VL-235B) are evaluated under both alternative judges, enabling a direct comparison of judge behaviour on identical responses. Table~\ref{tab:judge_agreement} reports overall pairwise Pearson correlation, within-1-point agreement rate, and ACC delta between GPT-5 and each alternative judge; Table~\ref{tab:judge_agreement_by_type} further stratifies by question type.

\begin{table}[htbp]
\centering
\small
\caption{Overall judge agreement statistics on the full \textsc{SynthDocBench} (200 reports, 1{,}788 questions). \textbf{$r$}: Pearson correlation between score sequences. \textbf{$w_1$}: fraction of responses where $|\text{score}_\text{GPT-5} - \text{score}_\text{alt}| \leq 1$. \textbf{$\Delta$ACC}: ACC(alt judge) $-$ ACC(GPT-5) at $\tau{=}6$.}
\label{tab:judge_agreement}
\begin{tabular}{@{}llccc@{}}
\toprule
\textbf{Candidate} & \textbf{Alt.\ Judge} & $r$ & $w_1$ & $\Delta$\textbf{ACC} \\
\midrule
GPT-4o           & Gemini-3.1-Pro    & 0.942 & 0.838 & $-$0.035 \\
Qwen3-VL-235B    & Gemini-3.1-Pro    & 0.960 & 0.893 & $-$0.010 \\
\midrule
GPT-4o           & Claude-Sonnet-4.5 & 0.881 & 0.654 & $+$0.135 \\
Qwen3-VL-235B    & Claude-Sonnet-4.5 & 0.916 & 0.723 & $+$0.156 \\
\bottomrule
\end{tabular}
\end{table}

\begin{table*}[htbp]
\centering
\small
\caption{Judge agreement stratified by question type on the full \textsc{SynthDocBench} (200 reports, 1{,}788 questions).
  \textbf{$r$}: Pearson correlation. \textbf{$w_1$}: within-1-point agreement. \textbf{$\Delta$ACC}: ACC(alt)~$-$~ACC(GPT-5) at $\tau{=}6$.
  Positive $\Delta$ACC means the alternative judge is more lenient than GPT-5.}
\label{tab:judge_agreement_by_type}
\resizebox{\textwidth}{!}{%
\begin{tabular}{@{}ll ccc ccc ccc ccc@{}}
\toprule
 & & \multicolumn{3}{c}{\textbf{Overall}} & \multicolumn{3}{c}{\textbf{Chart}} & \multicolumn{3}{c}{\textbf{Complex}} & \multicolumn{3}{c}{\textbf{Cross-Modal}} \\
\cmidrule(lr){3-5}\cmidrule(lr){6-8}\cmidrule(lr){9-11}\cmidrule(lr){12-14}
\textbf{Candidate} & \textbf{Alt.\ Judge} & $r$ & $w_1$ & $\Delta$\textbf{ACC} & $r$ & $w_1$ & $\Delta$\textbf{ACC} & $r$ & $w_1$ & $\Delta$\textbf{ACC} & $r$ & $w_1$ & $\Delta$\textbf{ACC} \\
\midrule
\multicolumn{14}{c}{\textit{Alt.\ Judge: Gemini-3.1-Pro}} \\
\midrule
GPT-4o        & Gemini & 0.942 & 0.838 & $-$0.035 & 0.953 & 0.801 & $-$0.030 & 0.915 & 0.869 & $-$0.054 & 0.950 & 0.845 & $-$0.020 \\
Qwen3-VL-235B & Gemini & 0.960 & 0.893 & $-$0.010 & 0.968 & 0.910 & $-$0.022 & 0.933 & 0.896 & $-$0.013 & 0.967 & 0.872 & $+$0.005 \\
\midrule
\multicolumn{14}{c}{\textit{Alt.\ Judge: Claude-Sonnet-4.5}} \\
\midrule
GPT-4o         & Claude & 0.881 & 0.654 & $+$0.135 & 0.913 & 0.647 & $+$0.022 & 0.833 & 0.610 & $+$0.271 & 0.895 & 0.705 & $+$0.113 \\
Qwen3-VL-235B  & Claude & 0.916 & 0.723 & $+$0.156 & 0.950 & 0.824 & $+$0.069 & 0.849 & 0.618 & $+$0.241 & 0.916 & 0.727 & $+$0.159 \\
\bottomrule
\end{tabular}%
}
\end{table*}

GPT-5 and Gemini-as-judge agree to within 3.5 ACC points ($r \geq 0.94$, $w_1 \geq 0.84$) across all tested candidates. Agreement is highest for chart-reading and cross-modal questions ($r \geq 0.95$) and slightly lower for complex multi-hop questions ($r \geq 0.91$), where scoring subjectivity is higher. These gaps are consistent across both candidate models and are well within the benchmark's bootstrap CI half-widths, confirming that model rankings are robust to judge choice regardless of question type. Claude-Sonnet-4.5 is a systematically lenient judge ($+$11--16 ACC points overall, rising to $+$24--27 pp on complex questions), consistent with the positivity bias documented for smaller LLM judges, and was excluded on this basis. The GPT-5/Gemini convergence addresses a potential vendor-conflict concern: Gemini-as-judge independently replicates GPT-5's ranking with a maximum ACC deviation of 3.5 points.

\paragraph{Ranking correlation with MMLongBench-Doc.}
To assess external validity, we compute the Spearman rank correlation between SynthDocBench overall ACC and published MMLongBench-Doc scores for the six models evaluated on both benchmarks. The correlation is $\rho = 0.657$ ($r = 0.683$), indicating moderate positive agreement: models that perform well on our benchmark tend to perform well on MMLongBench-Doc, but the rankings are not identical. Notably, GPT-4o ranks higher on MMLongBench-Doc (\#2) than on SynthDocBench (\#3), suggesting that our benchmark's emphasis on precise chart-value extraction and cross-modal alignment surfaces capabilities that MMLongBench-Doc's more heterogeneous question set partially masks. The imperfect correlation ($\rho < 1$) is itself evidence that SynthDocBench provides complementary diagnostic signal beyond existing benchmarks.

\paragraph{Qualitative alignment with real-document failures.}
Consistent with our benchmark's error taxonomy, the failure modes we identify in the controlled setting correspond to qualitatively similar patterns in MMLongBench-Doc: models that score lowest on our chart-reading subset also show the steepest degradation on MMLongBench-Doc's figure-heavy questions, and the cross-modal integration failures we isolate (where GPT-4o drops to 0.342 ACC) align with the documented difficulty of questions requiring evidence from non-adjacent pages in real documents. The key distinction is attribution: in real documents these factors co-vary and failures cannot be cleanly assigned to a single cause, whereas SynthDocBench's factorial design makes each failure mode individually observable.

\section{Bootstrap Confidence Intervals}
\label{app:bootstrap_ci}

Table~\ref{tab:bootstrap_ci} reports per-model bootstrap 95\% confidence intervals (2{,}000 resamples, seed 42) for ACC on each question subset; Table~\ref{tab:bootstrap_ci_diff} breaks these down by difficulty level. All pairwise ACC gaps between adjacent-ranked models exceed their combined CI half-widths, confirming the reliability of the rankings in Table~\ref{tab:main_results}.

\begin{table}[htbp]
\centering
\small
\caption{Bootstrap 95\% CI half-widths ($\pm$) for ACC at $\tau{=}6$, computed over 2{,}000 resamples (seed 42) on the full \textsc{SynthDocBench} (200 reports, 1{,}788 questions).}
\label{tab:bootstrap_ci}
\begin{tabular}{@{}lcccc@{}}
\toprule
\textbf{Model} & \textbf{Overall} & \textbf{Chart} & \textbf{Complex} & \textbf{Cross-Modal} \\
\midrule
Gemini-3.1-Pro    & $\pm$0.020 & $\pm$0.036 & $\pm$0.034 & $\pm$0.039 \\
GPT-4o            & $\pm$0.022 & $\pm$0.040 & $\pm$0.039 & $\pm$0.038 \\
Claude-Sonnet-4.5 & $\pm$0.022 & $\pm$0.041 & $\pm$0.037 & $\pm$0.034 \\
Qwen3-VL-235B     & $\pm$0.022 & $\pm$0.038 & $\pm$0.041 & $\pm$0.040 \\
InternVL3-78B     & $\pm$0.023 & $\pm$0.040 & $\pm$0.038 & $\pm$0.038 \\
Qwen2.5-VL-7B     & $\pm$0.013 & $\pm$0.029 & $\pm$0.008 & $\pm$0.021 \\
\bottomrule
\end{tabular}
\end{table}

\begin{table}[htbp]
\centering
\small
\caption{Bootstrap 95\% CI half-widths ($\pm$) for ACC at $\tau{=}6$ stratified by difficulty level, computed over 2{,}000 resamples (seed 42).}
\label{tab:bootstrap_ci_diff}
\begin{tabular}{@{}lccccc@{}}
\toprule
\textbf{Model} & \textbf{L1} & \textbf{L2} & \textbf{L3} & \textbf{L4} & \textbf{L5} \\
\midrule
Gemini-3.1-Pro    & $\pm$0.055 & $\pm$0.044 & $\pm$0.040 & $\pm$0.036 & $\pm$0.069 \\
GPT-4o            & $\pm$0.063 & $\pm$0.048 & $\pm$0.044 & $\pm$0.042 & $\pm$0.053 \\
Claude-Sonnet-4.5 & $\pm$0.070 & $\pm$0.044 & $\pm$0.044 & $\pm$0.042 & $\pm$0.051 \\
Qwen3-VL-235B     & $\pm$0.070 & $\pm$0.049 & $\pm$0.044 & $\pm$0.044 & $\pm$0.069 \\
InternVL3-78B     & $\pm$0.070 & $\pm$0.046 & $\pm$0.042 & $\pm$0.043 & $\pm$0.056 \\
Qwen2.5-VL-7B     & $\pm$0.018 & $\pm$0.031 & $\pm$0.032 & $\pm$0.015 & $\pm$0.008 \\
\bottomrule
\end{tabular}
\end{table}

\begin{figure}
  \centering
  \includegraphics[width=0.48\textwidth]{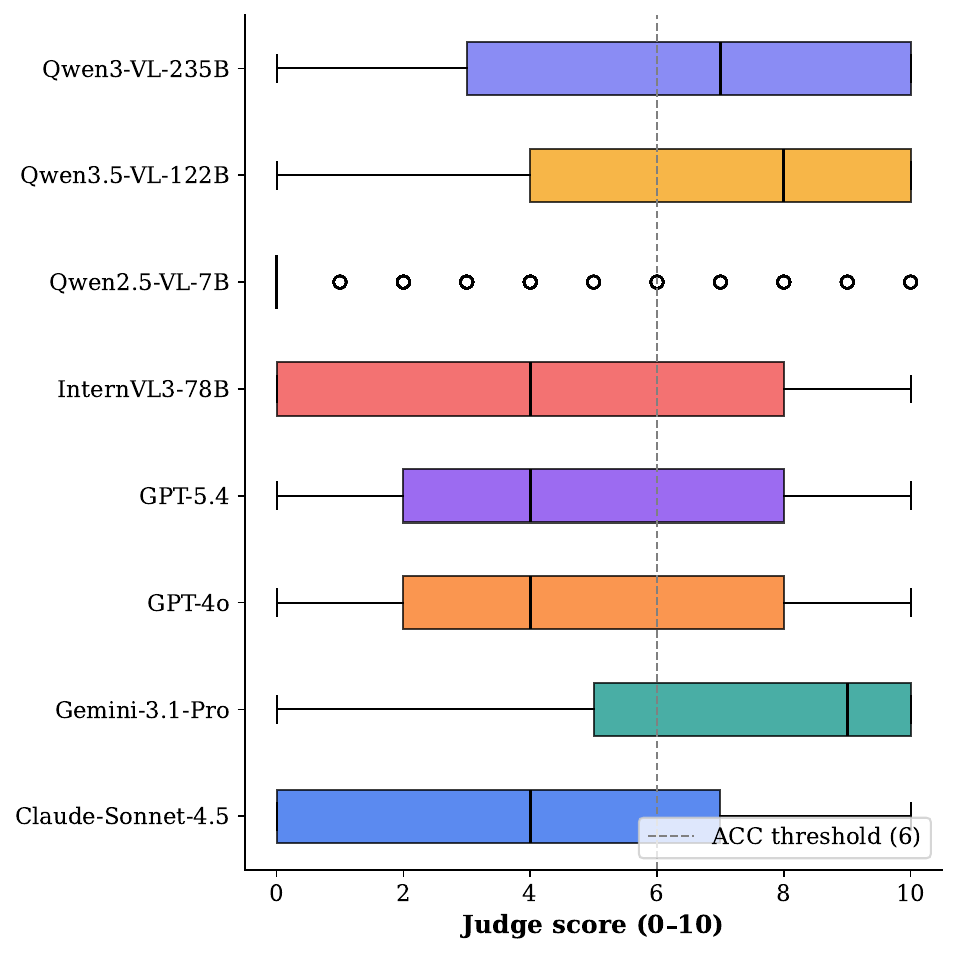}
  \caption{Distribution of judge scores (0--10) per model.}
  \label{fig:score_dist}
\end{figure}
\section{Score Distribution per Model}
\label{app:score_dist}

Figure~\ref{fig:score_dist} shows the distribution of raw judge scores (0--10) for each
model. A bimodal distribution (mass near 0--2 and near 8--10) indicates a model that
either fully recovers or fully misses an answer. A unimodal distribution centered at
6--7 indicates consistent but imprecise retrieval. Gemini-3.1-Pro is the only model
with a strong right-skewed distribution, while Qwen2.5-VL-7B is concentrated near 0.

\FloatBarrier
\section{Question Quality Control}
\label{app:qc}

We apply three layers of quality assurance to the generated questions.

\paragraph{Numeric recomputation.}
For every question whose ground-truth answer is a single numeric value (e.g.\ a chart data point, percentage, or count), we independently re-derive the answer from the structured metadata $\mathcal{M}$ used during generation, including raw \texttt{chart\_data} arrays and \texttt{key\_facts} tables.  Any question whose recomputed answer differs from the LLM-generated answer by more than a 5\% relative tolerance is flagged and either corrected or dropped.  This filter removed fewer than 2\% of chart-reading questions.

\paragraph{Automated consistency filtering.}
All 1,788 questions pass a rule-based filter checking: (1)~the question references an element present in $\mathcal{M}$; (2)~the ground-truth answer is non-empty and at least six tokens long; (3)~no exact-match overlap between question text and answer text exceeding 40\% of answer tokens (to screen out trivially answerable questions); and (4)~difficulty level $L$ is consistent with the number of inferential steps encoded in the \texttt{reasoning} field.

\paragraph{Manual review.}
A random sample of 100 questions (stratified by group type and difficulty level) was reviewed by two of the authors independently.  Reviewers rated each question on three dimensions: \emph{answerability} (is the answer determinable from the document?), \emph{answer correctness} (is the ground-truth answer correct?), and \emph{question clarity} (is the question unambiguous?).  Inter-rater agreement was $\kappa = 0.81$.  Overall acceptance rate across all three dimensions was 96\%.  The four rejected questions (two from \textit{complex} and two from \textit{cross-modal}) involved ambiguous referents that could not be resolved from the document alone; all four were removed from the final benchmark.

\FloatBarrier
\section{Positional Bias: Bar Chart}
\label{app:positional_bias}

\begin{figure}[htbp]
\centering

\begin{minipage}[t]{0.48\textwidth}
  \centering
  \includegraphics[width=\textwidth]{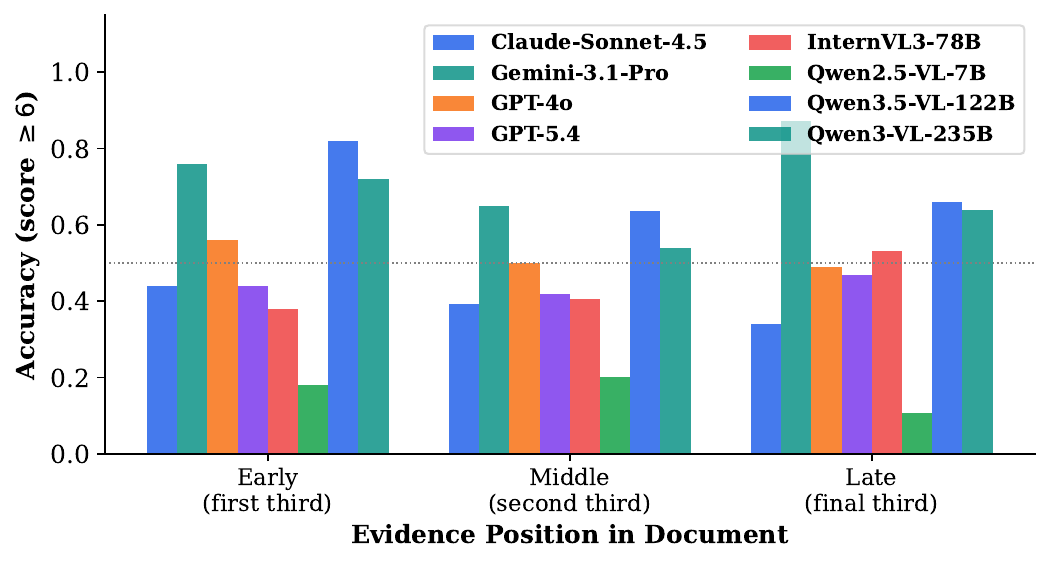}
  \caption{%
    Chart-reading ACC ($\tau{=}6$, judge: GPT-5) by evidence position bucket ($n{=}597$, 200 reports).
    Questions bucketed by relative chart position $p = k/K$ into equal thirds.
    Middle third is hardest for 4 of 6 models; Claude-Sonnet-4.5 steepest decline ($-$11.7~pp).
  }
  \label{fig:positional_bias}
\end{minipage}
\hfill
\begin{minipage}[t]{0.48\textwidth}
  \centering
  \includegraphics[width=\textwidth]{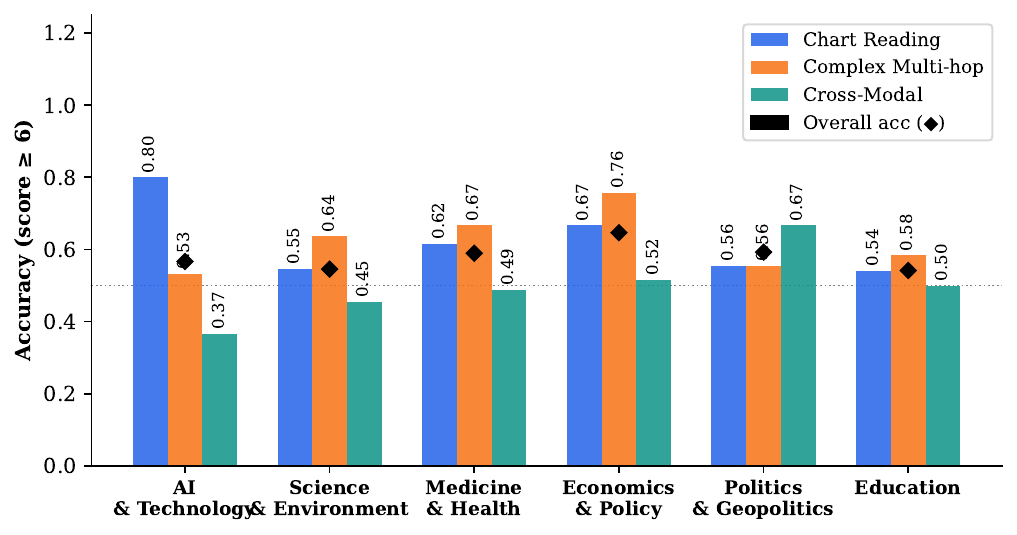}
  \caption{%
    Gemini-3.1-Pro accuracy (ACC, $\tau{=}6$) per topic domain, grouped by question type.
    Black diamonds = overall accuracy per domain. Dashed line at 0.5.
    Based on the 57 domain-annotated reports (513 questions).
  }
  \label{fig:domain_analysis}
\end{minipage}

\end{figure}

Figure~\ref{fig:positional_bias} visualises the ACC per position bucket from Table~\ref{tab:positional_bias} in the main text.

\FloatBarrier
\section{Domain Analysis}
\label{app:domain_analysis}

Figure~\ref{fig:domain_analysis} shows Gemini-3.1-Pro's accuracy stratified by six topic domains and three question types. Domain annotations cover 57 of the 200 reports (original curated set); the remaining 143 are not grouped by domain. Within the annotated subset, cross-modal accuracy falls 13--16 pp below chart-reading and complex in every domain; the gap is widest in AI \& Technology topics.

\FloatBarrier
\section{Question-Category Breakdown (Full Table)}
\label{app:category_table}

Table~\ref{tab:category_result_breakdown} provides the complete numerical values
visualised in Figure~\ref{fig:category_heatmap}.

\begin{table*}[htbp]
\centering
\footnotesize
\resizebox{\textwidth}{!}{%
\setlength{\tabcolsep}{4pt}
\begin{tabular}{@{}lcccccc@{}}
\toprule
\textbf{Category}
  & \textbf{Gemini-3.1} & \textbf{GPT-4o}
  & \textbf{Claude-S-4.5} & \textbf{Qwen3-VL} & \textbf{InternVL3}
  & \textbf{Qwen2.5-7B} \\
\midrule
\rowcolor{gray!15} \multicolumn{7}{c}{\textit{Chart Reading}} \\
\midrule
Value reading       & \textbf{0.784} & 0.271 & 0.382 & 0.648 & 0.437 & 0.015 \\
Comparison          & \textbf{0.759} & 0.377 & 0.266 & 0.623 & 0.327 & 0.075 \\
Trend / pattern     & 0.734          & \textbf{0.724} & 0.412 & 0.653 & 0.603 & 0.397 \\
\midrule
\rowcolor{gray!15} \multicolumn{7}{c}{\textit{Cross-Modal}} \\
\midrule
Verify with chart   & \textbf{0.616} & 0.399 & 0.278 & 0.480 & 0.318 & 0.045 \\
Integrate sources   & \textbf{0.692} & 0.298 & 0.260 & 0.545 & 0.293 & 0.030 \\
Compare repr.       & \textbf{0.576} & 0.328 & 0.213 & 0.485 & 0.278 & 0.126 \\
\midrule
\rowcolor{gray!15} \multicolumn{7}{c}{\textit{Complex Reasoning}} \\
\midrule
Historical/timeline & \textbf{0.910} & 0.553 & 0.508 & 0.784 & 0.598 & 0.015 \\
Technical           & \textbf{0.643} & 0.111 & 0.137 & 0.427 & 0.156 & 0.000 \\
Impact / reception  & \textbf{0.814} & 0.417 & 0.364 & 0.623 & 0.437 & 0.020 \\
\bottomrule
\end{tabular}
}
\caption{%
  ACC ($\tau{=}6$) by fine-grained question category (numerical companion to
  Figure~\ref{fig:category_heatmap}).
  Best per row in \textbf{bold}.
}
\label{tab:category_result_breakdown}
\end{table*}

\section{Performance by Question Subset}
\label{app:subset_fig}

\begin{figure}
  \centering
  \includegraphics[width=0.53\textwidth]{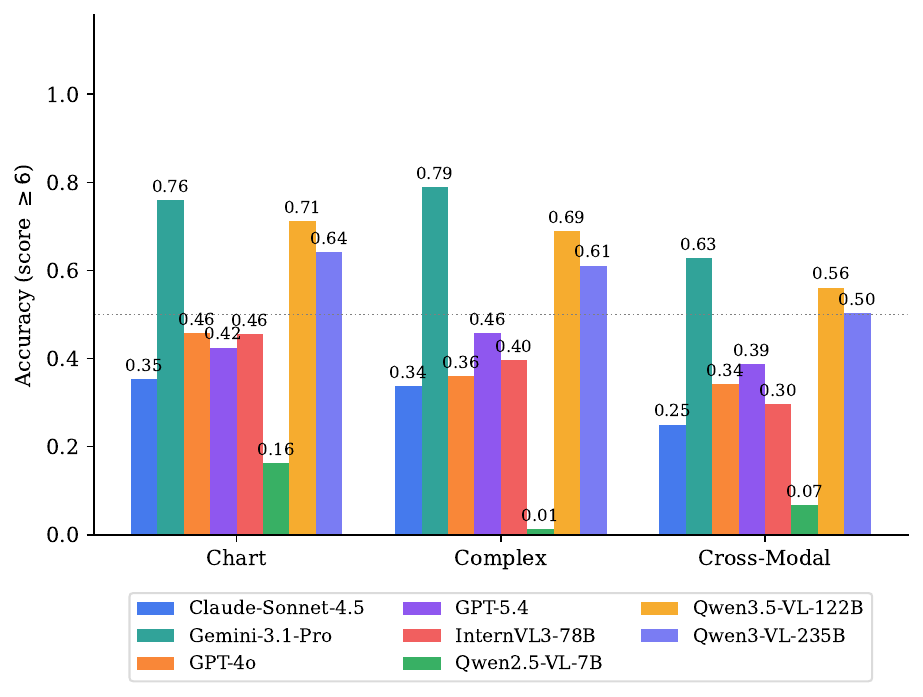}
  \caption{%
    ACC ($\tau{=}6$) by question subset. Cross-modal questions are consistently
    hardest across all models, confirming modality alignment as the primary bottleneck.
  }
  \label{fig:perf_by_subset}
\end{figure}

Figure~\ref{fig:perf_by_subset} reports ACC ($\tau{=}6$) broken down by question subset
(chart-reading, complex, cross-modal) for all six models.
Cross-modal integration is consistently the hardest subset across all models,
confirming modality alignment as the primary bottleneck in long-context document understanding.

\FloatBarrier
\section{Question Category Taxonomy}
\label{app:question_taxonomy}

Table~\ref{tab:question_taxonomy} defines all nine fine-grained question categories
used in the category-level breakdown of Table~\ref{tab:category_result_breakdown}.
Each category belongs to one of the three question subsets and targets a distinct
reasoning capability. The difficulty label $L1$--$L5$ is orthogonal to category:
any category can appear at any difficulty level depending on the complexity of the
evidence chain.

\begin{table}[htbp!]
\centering
\small
\resizebox{\textwidth}{!}{%
\begin{tabular}{@{} l c c l @{}}
\toprule
\textbf{Level} & \textbf{Modality} & \textbf{Operation} & \textbf{Example} \\
\midrule
L1: Direct lookup         & Chart / Table    & Read-off  & \texttt{What is the value for \emph{X} in chart \emph{Y}?} \\
L2: Comparison            & Chart / Table    & Compare   & \texttt{Which is larger, \emph{A} or \emph{B}?} \\
L3: Aggregation           & Chart / Table    & Compute   & \texttt{What is the total or average of $\ldots$?} \\
L4: Domain reasoning      & Chart + Text     & Infer     & \texttt{Why does \emph{X} outperform \emph{Y} given the context?} \\
L5: Visual interpretation & Chart only       & Interpret & \texttt{What does colour \emph{C} encode in chart \emph{Y}?} \\
\bottomrule
\end{tabular}
}
\caption{Difficulty taxonomy for \synthdocbench{}}
\label{tab:difficulty-levels}
\end{table}

\begin{table}[htbp]
\centering
\footnotesize
\caption{%
  Fine-grained question category definitions for \synthdocbench{}.
  Categories are grouped by subset; all nine are mutually exclusive within their subset.
}
\label{tab:question_taxonomy}
\begin{tabular}{@{}lp{10.5cm}@{}}
\toprule
\textbf{Category} & \textbf{Definition} \\
\midrule
\rowcolor{gray!15} \multicolumn{2}{c}{\textit{Chart Reading}} \\
\midrule
Value Reading &
  Extract one or more exact numerical values from a single chart or table element
  (e.g., read a bar height, a cell in a table, or a point on a line). No cross-source
  reasoning is required; the answer is localized to a single visual element. \\[4pt]
Comparison &
  Compare two or more values within the same chart or table to identify the larger,
  smaller, or closest item, or to compute a difference or ratio. The answer requires
  reading at least two elements from the same visual. \\[4pt]
Trend / Pattern &
  Describe the directional behavior, distribution shape, or temporal pattern visible
  in a chart (e.g., ``increasing trend,'' ``bimodal distribution,'' ``peaks in Q3'').
  Exact numerical precision is secondary to correctly characterizing the overall pattern. \\
\midrule
\rowcolor{gray!15} \multicolumn{2}{c}{\textit{Cross-Modal}} \\
\midrule
Verify with Chart &
  A specific quantitative claim appears in the document text. The model must locate
  the corresponding chart and confirm, refute, or quantify the claim using visual evidence. \\[4pt]
Integrate Sources &
  Answer a question whose complete response requires combining a textual claim with a
  specific numerical value readable only from an associated chart. Neither source alone
  is sufficient; the model must execute a four-step pipeline: locate claim~$\to$ find
  chart~$\to$ parse value~$\to$ synthesize. \\[4pt]
Compare Representations &
  The same phenomenon is described both in text (e.g., a percentage trend) and in a
  chart (e.g., a line plot). The model must reconcile or contrast these two representations,
  identifying any discrepancy or providing a richer joint characterization. \\
\midrule
\rowcolor{gray!15} \multicolumn{2}{c}{\textit{Complex Reasoning}} \\
\midrule
Historical / Timeline &
  Synthesize facts distributed across multiple sections or time periods to construct a
  causal or chronological narrative (e.g., sequence of policy changes, evolution of a
  metric over decades). Requires integrating at least two temporally separated evidence units. \\[4pt]
Technical / Quantitative &
  Apply domain-specific knowledge or multi-step arithmetic to evidence recovered from
  the document (e.g., derive a compound growth rate, convert units, or evaluate a
  quantitative tradeoff between two technical approaches). \\[4pt]
Impact / Reception &
  Assess the downstream effects, societal implications, or critical reception of
  a phenomenon discussed in the document. Requires synthesizing evaluative language
  from multiple sections rather than extracting a single localized fact. \\
\bottomrule
\end{tabular}
\end{table}

\FloatBarrier
\section{Additional Implementation Details}
\label{app:implementation_details}

\begin{figure}
  \centering
  \includegraphics[width=0.70\linewidth]{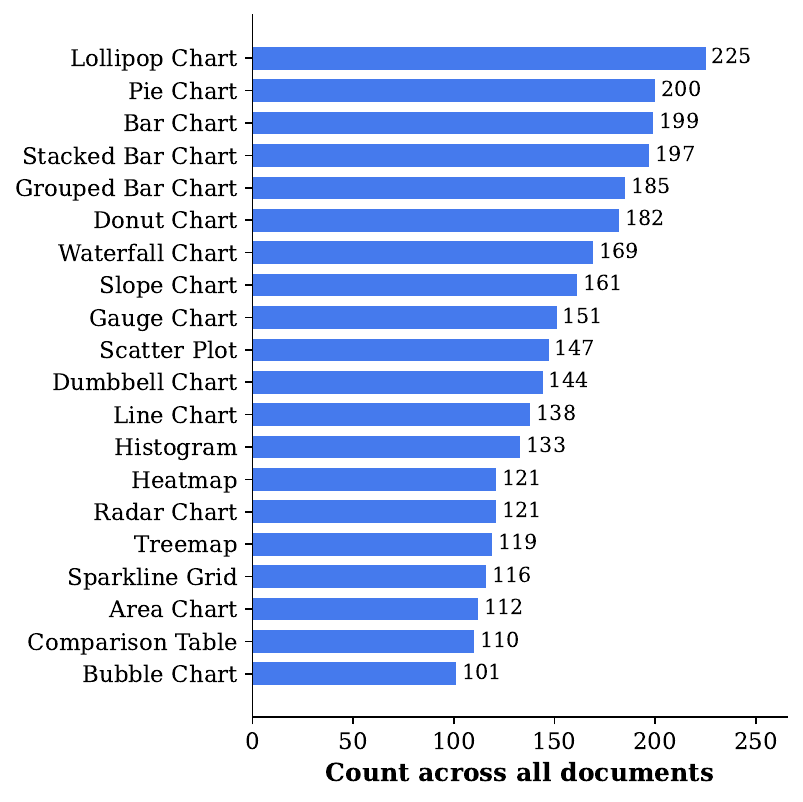}
  \caption{Distribution of chart types (top 20 shown). The corpus covers
    24 distinct types spanning common (bar, line, scatter) and specialized
    (dumbbell, sankey, lollipop) forms.}
  \label{fig:chart_type_dist}
\end{figure}

This appendix summarizes implementation details that support reproducibility but are not
strictly necessary for understanding the main methodological contributions. We include
them here to document the design space of document layouts, the structured schema used to
encode visualization semantics, the serialized format of generated QA items, and the
rendering constraints imposed by the evaluation harness.

\subsection{Notation Glossary}
\label{app:notation}

For readability, the main text introduces notation only when needed. Table~\ref{tab:notation_appendix}
collects the full symbol glossary in one place. The notation spans the three benchmark
stages: synthetic document generation, structured question generation, and image-based
evaluation.

\begin{table}[htbp]
\centering
\small
\caption{Notation used in the \textsc{SynthDocBench} methodology.}
\label{tab:notation_appendix}
\begin{tabular}{clp{10cm}}
\toprule
\textbf{Symbol} & \textbf{Type} & \textbf{Definition} \\
\midrule
$\tau$        & string     & Topic seed used to initialize document generation \\
$\mathcal{A}$ & set        & Layout archetype space \\
$a \sim P_{\mathcal{A}}(\tau)$ & sample & Archetype sampled from the topic-conditioned archetype distribution \\
$\mathcal{D}$ & document   & Final rendered document, represented as HTML and exported PDF \\
$\mathcal{V}_k$ & object   & Structured metadata object associated with the $k$-th visualization in $\mathcal{D}$ \\
$\mathcal{T}_k$ & object   & Structured metadata object associated with the $k$-th table in $\mathcal{D}$ \\
$\mathcal{M}$ & manifest   & Document-level QA manifest aggregating all $\mathcal{V}_k$ and $\mathcal{T}_k$ \\
$q$           & string     & Natural-language evaluation question \\
$a^*$         & string     & Reference answer derived deterministically from $\mathcal{M}$ \\
$\hat{a}$     & string     & Model-predicted answer for $q$ from rendered page image(s) of $\mathcal{D}$ \\
$f_\theta$    & model      & Candidate vision-language model parameterized by $\theta$ \\
$\mathcal{I}$ & sequence   & Rendered page-image sequence supplied to the evaluated model \\
$\ell$        & integer    & Difficulty level associated with a question \\
$\mathcal{J}$ & model      & Judge model used to score $(\hat{a}, a^*)$ \\
\bottomrule
\end{tabular}
\end{table}

\subsection{Layout Archetypes}
\label{app:archetypes}

The report-generation pipeline uses a small but diverse inventory of layout archetypes to
vary document structure, rhetorical style, and chart placement strategy. These archetypes
control the global page grammar of a report, including whether visualizations are embedded
inline, arranged in grids, or surfaced as full-width elements. They also affect the types
of auxiliary components that can appear, such as metric strips, pull quotes, or timeline
bands. Table~\ref{tab:archetypes_appendix} lists the archetypes used in
\textsc{SynthDocBench} together with their distinctive design characteristics.

\begin{table}[htbp]
\centering
\small
\caption{Layout archetypes used in \textsc{SynthDocBench}.}
\label{tab:archetypes_appendix}
\begin{tabular}{lllcp{5.5cm}}
\toprule
\textbf{Archetype} & \textbf{Viz layout} & \textbf{Font mood} & \textbf{Max vizzes} & \textbf{Distinctive features} \\
\midrule
Magazine    & Hero + smaller   & Editorial  & 3 & 90vh hero, drop caps, gradient overlays, pull quotes \\
Dashboard   & $2\times$ grid   & Technical  & 4 & KPI strip, metric cards, compact chart grids \\
Academic    & Inline           & Scholarly  & 2 & Abstract box, numbered sections, formal tables \\
Editorial   & Breakout column  & Narrative  & 2 & Reading progress cues, chapter markers, block quotes \\
Infographic & Full-width stack & Bold       & 4 & Icon strips, timeline bands, data callouts \\
Brutalist   & Full bleed       & Raw        & 2 & Thick borders, monospace, oversized numbers, stamp labels \\
\bottomrule
\end{tabular}
\end{table}

\subsection{Visualization Metadata Schema}
\label{app:viz_schema}

A key property of the benchmark is that every generated visualization is accompanied by a
structured metadata object that records its semantic content independently of the rendered
pixels. This metadata is used downstream for deterministic QA generation and for
consistency validation between rendered charts and their underlying values. The schema is
designed to capture both data semantics (e.g., chart values, axes, and derived insights)
and presentation-level attributes (e.g., legend presence, color encoding, and highlighted
items). Table~\ref{tab:viz_schema_appendix} summarizes the fields stored for each chart.

\begin{table}[!htbp]
\centering
\small
\caption{Visualization metadata schema used by the report-generation pipeline.}
\label{tab:viz_schema_appendix}
\begin{tabular}{llp{10cm}}
\toprule
\textbf{Field} & \textbf{Type} & \textbf{Contents} \\
\midrule
\texttt{vizId}            & string  & Unique identifier of the form \texttt{viz-\{section\}-\{index\}} \\
\texttt{chartType}        & enum    & Chart type associated with the rendered visualization \\
\texttt{title}            & string  & Descriptive chart title \\
\texttt{axes.x / axes.y}  & object  & Axis label, type, optional unit, and optional range \\
\texttt{data}             & array   & $[$ \{label, value, category, metadata\} $]$ --- one entry per rendered mark \\
\texttt{insights}         & array   & Typed facts such as maximum, minimum, comparison, gap, trend, outlier, or proportion \\
\texttt{visualProperties} & object  & colorEncoding, colorMap, sortOrder, hasLegend, hasGridlines, highlightedItems, annotationCount \\
\texttt{sourceContext}    & string  & Attribution to the source passage supplying the underlying values \\
\bottomrule
\end{tabular}
\end{table}

\subsection{Question Output Schema}
\label{app:qa_schema}

Generated QA items are serialized in a structured format so that evaluation can be
performed not only at the answer level but also at the level of question family,
difficulty, and supporting evidence. In addition to the surface question and reference
answer, the output schema stores evidence traces such as required facts, chart indices,
and required data points, which are useful for debugging, validation, and downstream
error analysis. Table~\ref{tab:qa_output_appendix} describes the fields included in each
serialized QA record.

\begin{table}[htbp]
\centering
\small
\caption{Structured output schema for generated question-answer pairs.}
\label{tab:qa_output_appendix}
\begin{tabular}{llp{6.8cm}}
\toprule
\textbf{Field} & \textbf{Question types} & \textbf{Contents} \\
\midrule
\texttt{question}           & All       & Natural-language question string \\
\texttt{answer}             & All       & Reference answer \\
\texttt{question\_type}     & All       & Question family/type label \\
\texttt{difficulty}         & All       & Difficulty level $L1$--$L5$ \\
\texttt{required\_facts}    & Multi-hop & Indices into the extracted evidence list \\
\texttt{required\_facts\_text} & Multi-hop & Verbatim supporting evidence \\
\texttt{fact\_sources}      & Multi-hop & Source section for each supporting fact \\
\texttt{reasoning}          & All       & Stepwise evidence-to-answer derivation \\
\texttt{chart\_index}       & Visual, Cross-modal & Index of the referenced chart \\
\texttt{chart\_title}       & Visual, Cross-modal & Title of the referenced chart \\
\texttt{required\_data\_points} & Visual, Cross-modal & Specific \{label, value\} pairs required for the answer \\
\texttt{category}           & Visual, Cross-modal & Fine-grained category label \\
\bottomrule
\end{tabular}
\end{table}

\subsection{Rendering Hyperparameters}
\label{app:rendering_details}

The evaluation harness operates on rendered page images rather than HTML or PDF source,
so rendering choices directly affect the visual evidence available to the model. The
hyperparameters listed in Table~\ref{tab:rendering_appendix} define the image
construction process, including rasterization resolution, maximum page budget, batching
of pages into concatenated strips, and compression constraints required to satisfy API
limits across providers. These values therefore influence both evaluation efficiency and
the effective difficulty of the visual inference problem.

\begin{table}[htbp]
\centering
\small
\caption{Rendering and compression hyperparameters.}
\label{tab:rendering_appendix}
\begin{tabular}{llr}
\toprule
\textbf{Parameter} & \textbf{Description} & \textbf{Value} \\
\midrule
\texttt{RESOLUTION}      & PDF rasterization DPI                     & 144 \\
\texttt{MAX\_PAGES}      & Maximum pages extracted per PDF           & 120 \\
\texttt{CONCAT\_NUM}     & Pages per concatenated image strip        & 5   \\
\texttt{COLUMN\_NUM}     & Columns per concatenated image            & 1   \\
\texttt{MAX\_IMG\_BYTES} & Maximum compressed image size             & 4~MB \\
\texttt{MAX\_DIMENSION}  & Maximum pixel dimension (longest side)    & 7{,}900~px \\
\texttt{ACC\_THRESHOLD}  & Minimum judge score counted as correct    & 6   \\
\bottomrule
\end{tabular}
\end{table}

\subsection{Inference and Judge Configuration}
\label{app:inference_details}

Table~\ref{tab:inference_appendix} summarizes the fixed inference configuration applied
to both candidate models and the judge. All models are queried at temperature~0 to
ensure fully deterministic outputs across runs.

\begin{table}[htbp]
\centering
\small
\caption{Inference configuration for candidate models and the judge.}
\label{tab:inference_appendix}
\begin{tabular}{llp{7cm}}
\toprule
\textbf{Parameter} & \textbf{Value} & \textbf{Notes} \\
\midrule
Judge model        & GPT-5          & Fixed across all experiments \\
Temperature        & 0              & Applied to both candidates and judge \\
Candidate response & 2--4 sentences & Enforced via system prompt \\
Judge format       & JSON only      & \texttt{\{"score": <0--10>, ...\}} \\
Parse failure      & Score $= -1$   & Excluded from $Q_{\mathrm{valid}}$ \\
\bottomrule
\end{tabular}
\end{table}

\subsection{Prompts}
\label{app:prompts}

All prompts are shown verbatim as supplied to the model APIs.
Text in \textit{italics} denotes fields populated at runtime.

\paragraph{Candidate model — default (concise) prompt.}

\begin{promptbox}[\textsc{System}]
\ttfamily\small
You are an expert analyst reviewing a PDF report.\\
Answer the question using ONLY what is visible in the provided pages.\\
Be concise and direct --- 2--4 sentences maximum.\\
Include specific numbers, dates, or names only if directly relevant.\\
Do not restate the question. Do not explain your reasoning. Just answer.\\
If the answer is not in the document, say: ``Not found in the document.''
\end{promptbox}

\begin{promptbox}[\textsc{User}]
\ttfamily\small
\textit{[Rendered page images supplied as vision input]}\\[3pt]
Question: \textit{\{question\}}
\end{promptbox}

\paragraph{Candidate model — chain-of-thought (CoT) prompt.}

\begin{promptbox}[\textsc{System}]
\ttfamily\small
You are an expert analyst reviewing a PDF report.\\
Answer the question using ONLY what is visible in the provided pages.\\
Think step by step: first identify the relevant chart or section, then extract\\
the key facts, then synthesize your final answer.\\
Include specific numbers, dates, or names where relevant.\\
If the answer is not in the document, say: ``Not found in the document.''
\end{promptbox}

\begin{promptbox}[\textsc{User}]
\ttfamily\small
\textit{[Rendered page images supplied as vision input]}\\[3pt]
Question: \textit{\{question\}}
\end{promptbox}

\paragraph{Candidate model — no system prompt baseline.}
In the \textit{none} condition no system prompt is supplied.
The user turn contains only the rendered page images followed by the question,
identical to the user turns above.

\paragraph{Judge model prompt.}

\begin{promptbox}[\textsc{System}]
\ttfamily\small
You are a factual accuracy evaluator.\\
Score the CANDIDATE ANSWER against the REFERENCE ANSWER for a document-based Q\&A task.\\[5pt]
Rubric (0--10):\\
\phantom{xx}10\ = all key facts present and correct\\
\phantom{xx}8--9 = mostly correct, minor gap or imprecision\\
\phantom{xx}6--7 = core correct, some detail missing\\
\phantom{xx}4--5 = partial, significant gaps or errors\\
\phantom{xx}2--3 = mostly wrong or missing\\
\phantom{xx}0--1 = incorrect or hallucinated\\[5pt]
Respond with ONLY this JSON (no markdown, no extra text):\\
\{"score": <0--10>, "judgement": "<2--3 sentences on accuracy>"\}
\end{promptbox}

\begin{promptbox}[\textsc{User}]
\ttfamily\small
QUESTION: \textit{\{question\}}\\[4pt]
REFERENCE ANSWER: \textit{\{gt\_answer\}}\\[4pt]
CANDIDATE ANSWER: \textit{\{candidate\_answer\}}\\[4pt]
Score 0--10 and explain. Return JSON only.
\end{promptbox}

\FloatBarrier
\section{Hard Failure Error Category Definitions}
\label{app:error_categories}

\begin{table}[htbp]
\centering
\footnotesize
\setlength{\tabcolsep}{6pt}
\begin{tabular}{@{}lcccc@{}}
\toprule
\textbf{Subset} & \textbf{Total} & \textbf{Ch.} & \textbf{Cx.} & \textbf{XM.} \\
\midrule
Hard failures (all 6 $\leq 3$) & \textbf{109} & \textbf{37} & \textbf{14} & \textbf{58} \\
\bottomrule
\end{tabular}
\caption{Hard failure counts by question subset (all six models score $\leq 3$).
  \textbf{Ch.}~= chart-reading; \textbf{Cx.}~= complex; \textbf{XM.}~= cross-modal.}
\label{tab:hard_failure_counts}
\end{table}

Table~\ref{tab:error_categories_app} defines the seven mutually exclusive error
categories used to classify hard failures in Section~\ref{sec:error_analysis}.
Each category was assigned by GPT-5 based on the question, ground-truth answer,
and all model responses.
The per-category count breakdown is in Table~\ref{tab:hard_failure_counts}.

\begin{table}[htbp]
\centering
\footnotesize
\caption{%
  Error category definitions for the hard failure analysis
  (Section~\ref{sec:error_analysis}).
  Categories are mutually exclusive; each failure is assigned exactly one.
}
\label{tab:error_categories_app}
\begin{tabular}{@{}lp{9cm}@{}}
\toprule
\textbf{Category} & \textbf{Definition} \\
\midrule
Visual Hallucination &
  The model confidently reads a wrong value from the chart.
  The figure is located and the relevant element identified, but the
  pixel-level value extraction is incorrect (e.g., reports 1{,}000~ppm instead of 30~ppm). \\[4pt]
Figure Not Found &
  The model cannot locate the referenced figure in the document.
  It either returns ``not found in the document'' or retrieves a different, unrelated figure. \\[4pt]
Precision Error &
  The model extracts the correct chart element but with the wrong exact value, scale,
  or unit (e.g., reads 29\% instead of 27.5\%, or reports billions instead of millions). \\[4pt]
Incomplete Retrieval &
  The model retrieves part of the required evidence but misses one or more key facts
  distributed across the document, producing an answer that is partially correct but
  substantively incomplete. \\[4pt]
Cross-Modal Grounding &
  Both the textual claim and the relevant chart are correctly identified in isolation,
  but the model fails to perform the required quantitative alignment or comparison
  between the two modalities. \\[4pt]
Reasoning Error &
  The necessary evidence is correctly retrieved from the document, but the model draws
  a wrong conclusion through a logical or arithmetic mistake in the final inference step. \\[4pt]
Question / Annotation Ambiguity &
  The question or ground-truth answer is ambiguous, under-specified, or requires
  information that is not recoverable from the visible document pages. \\
\bottomrule
\end{tabular}
\end{table}

\FloatBarrier
\section{Ablation Study Details}
\label{app:ablation_details}

This section documents the exact experimental configuration used for each
ablation study.
All ablations use Gemini-3.1-Pro as the candidate model and GPT-5 as judge
unless otherwise stated. Each condition is evaluated on a representative
full 200-report benchmark (1,788 questions), and the same judge prompt is used throughout.

\subsection{Page Concatenation Ablation}
\label{app:abl_concat}

The evaluation harness renders each PDF page as a raster image and groups
consecutive pages into a single vertical image strip before passing them to
the model. The \texttt{concat-num} hyperparameter controls how many pages
are stacked per strip.

\begin{table}[htbp]
\centering
\begin{minipage}[t]{0.48\textwidth}
\centering
\resizebox{\textwidth}{!}{%
\begin{tabular}{@{}lll@{}}
\toprule
\textbf{Parameter} & \textbf{Value} & \textbf{Note} \\
\midrule
\texttt{--concat-num} & 1, 2, 5, 10 & Pages per image strip \\
\texttt{--resolution} & 144 & DPI fixed \\
\texttt{--prompt-style} & default & Concise system prompt \\
Model & gemini-3.1-pro & Fixed candidate \\
Total images/doc & $\lceil$pages / concat-num$\rceil$ & Varies \\
\bottomrule
\end{tabular}%
}
\caption{Concat-num ablation configuration.}
\label{tab:ablation_concat}
\end{minipage}
\hfill
\begin{minipage}[t]{0.48\textwidth}
\centering
\resizebox{\textwidth}{!}{%
\begin{tabular}{@{}lll@{}}
\toprule
\textbf{Parameter} & \textbf{Value} & \textbf{Note} \\
\midrule
\texttt{--resolution} & 72, 144, 216 & PDF rasterization DPI \\
\texttt{--concat-num} & 5 & Fixed (default) \\
\texttt{--prompt-style} & default & Fixed \\
Max bytes/image & 4\,MB & JPEG recompressed if exceeded \\
Max dimension & 7{,}900\,px & Downscaled if exceeded \\
\bottomrule
\end{tabular}%
}
\caption{DPI ablation configuration.}
\label{tab:ablation_dpi}
\end{minipage}
\end{table}

\noindent At 72\,DPI the resulting strip is typically around 1{,}200\,px tall;
at 144\,DPI around 2{,}400\,px; at 216\,DPI around 3{,}600\,px before any
compression rescaling. At \texttt{concat-num=1} each page is sent individually, maximising resolution per image but fragmenting document context across many turns.
At \texttt{concat-num=10}, up to ten pages are merged into one tall strip,
providing wider context at reduced per-line resolution.
The default \texttt{concat-num=5} balances these trade-offs.

\subsection{Rendering Resolution Ablation}
\label{app:abl_dpi}

PDF rasterization converts each page to a pixel image at the specified DPI.
Higher DPI produces finer text and chart detail but larger file sizes
that may trigger per-image byte-limit compression.

\subsection{Prompting Strategy Ablation}
\label{app:abl_prompt}

Across all three models, the no-prompt condition achieves the highest complex-question ACC
(Gemini $+$5.3 pts, GPT-4o $+$5.9 pts, Claude $+$17.0 pts over default),
with Claude's complex ACC jumping from 0.380 to 0.550, suggesting the concise-answer
framing actively constrains multi-step reasoning.
CoT yields modest overall gains (Gemini: $+$1.6 pts; Claude: $+$3.0 pts) but reduces
GPT-4o chart-reading ACC (0.520$\to$0.468), where longer outputs introduce hallucinated
intermediate steps.
Chart-reading and cross-modal ACC vary by at most 5 points across all strategies,
confirming these subsets are driven by visual perception rather than output formatting.

\begin{table}[htbp]
\centering
\begin{minipage}[t]{0.48\textwidth}
\centering
\resizebox{\textwidth}{!}{%
\begin{tabular}{@{}lll@{}}
\toprule
\textbf{Parameter} & \textbf{Values} & \textbf{Note} \\
\midrule
Judge models & GPT-5, Gemini, Claude & Same rubric/JSON \\
Candidate models & GPT-4o, Qwen3, InternVL3 & Pre-scored answers \\
ACC threshold $\tau$ & 4, 5, 6, 7, 8 & GPT-5 judge only \\
Temperature & 0 & All judges \\
\bottomrule
\end{tabular}%
}
\caption{Judge sensitivity ablation configuration.}
\label{tab:ablation_config}
\end{minipage}
\hfill
\begin{minipage}[t]{0.48\textwidth}
\centering
\resizebox{\textwidth}{!}{%
\begin{tabular}{@{}lll@{}}
\toprule
\textbf{Style} & \textbf{System prompt} & \textbf{Description} \\
\midrule
\texttt{default} & Concise (App.~\ref{app:prompts}) & 2--4 sentence answer \\
\texttt{cot}     & CoT (App.~\ref{app:prompts})     & Step-by-step reasoning \\
\texttt{none}    & None                              & No system prompt \\
\bottomrule
\end{tabular}%
}
\caption{Prompting ablation configuration.}
\label{tab:ablation_prompt}
\end{minipage}
\end{table}

\subsection{Judge Sensitivity Ablation}
\label{app:abl_judge}
\label{app:judge_sensitivity}

Judge sensitivity results are in Table~\ref{app:prompts}.
Across all three models, the no-prompt condition achieves the highest complex-question ACC
(Gemini $+$5.3 pts, GPT-4o $+$5.9 pts, Claude $+$17.0 pts over default),
with Claude's complex ACC jumping from 0.380 to 0.550, suggesting the concise-answer
framing actively constrains multi-step reasoning.
CoT yields modest overall gains (Gemini: $+$1.6 pts; Claude: $+$3.0 pts) but reduces
GPT-4o chart-reading ACC (0.520$\to$0.468), where longer outputs introduce hallucinated
intermediate steps.
Chart-reading and cross-modal ACC vary by at most 5 points across all strategies,
confirming these subsets are driven by visual perception rather than output formatting.

\begin{table}[h]
\centering
\footnotesize
\label{tab:ablation_judge}
\begin{tabular}{@{}llccc@{}}
\toprule
& & \textbf{GPT-4o} & \textbf{Qwen3-VL} & \textbf{InternVL3} \\
\midrule
\multirow{3}{*}{\textit{Judge model}}
  & GPT-5 (default)             & 0.434 & 0.535 & 0.354 \\
  & Gemini-3.1-Pro              & 0.417 & 0.522 & 0.363 \\
  & Claude-Sonnet-4.5           & 0.607 & 0.705 & 0.573 \\
\midrule
\multirow{5}{*}{\textit{Threshold $\tau$}}
  & $\tau=4$           & 0.627 & 0.697 & 0.568 \\
  & $\tau=5$           & 0.483 & 0.575 & 0.398 \\
  & $\tau=6$ (default) & 0.434 & 0.535 & 0.354 \\
  & $\tau=7$           & 0.360 & 0.451 & 0.311 \\
  & $\tau=8$           & 0.299 & 0.413 & 0.269 \\
\bottomrule
\end{tabular}
\caption{%
  Judge sensitivity results.
  \textit{Top:} ACC for three candidates under three judge models.
  \textit{Bottom:} ACC under GPT-5 at thresholds $\tau \in \{4,5,6,7,8\}$.
  Rankings are fully preserved across all conditions.
}
\end{table}

\section{Qualitative Model Response Examples}
\label{app:qualitative}


We present four representative examples from \synthdocbench{} spanning the three
question subsets and dominant failure modes from Section~\ref{sec:error_analysis}.
Each example shows (i)~a zoomed-in crop of the relevant evidence page,
(ii)~the question and ground-truth answer, and (iii)~the annotated response from
a representative subset of evaluated models, where \textcolor{red}{\textbf{green}} marks correct facts and
\textcolor{red}{\textbf{red}} marks hallucinated or wrong values.
\definecolor{mygreen}{RGB}{34, 139, 34}
\newcommand{\scorehi}[1]{\colorbox{lbgreen}{\small\bfseries\texttt{#1}}}
\newcommand{\scoremid}[1]{\colorbox{localyellow}{\small\bfseries\texttt{#1}}}
\newcommand{\scorelo}[1]{\colorbox{red!18}{\small\bfseries\texttt{#1}}}
\newcommand{\cor}[1]{\textcolor{mygreen}{\textbf{#1}}}
\newcommand{\wrg}[1]{\textcolor{red}{\textbf{#1}}}

\clearpage
\subsection*{Example 1 — Chart Reading L1 \hfill \textit{Mixed: Value Reading from Horizontal Bar}}
\noindent\rule{\linewidth}{0.4pt}
\begin{minipage}[t]{0.40\textwidth}
\vspace{0pt}
\centering
\includegraphics[width=\linewidth]{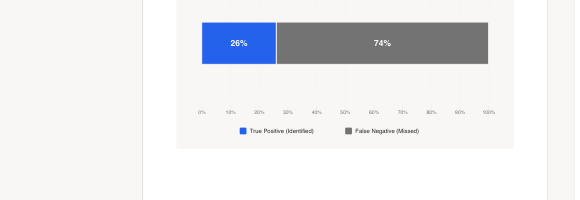}\\[3pt]
\footnotesize\textit{Figure~14 from the document (p.19) --- OpenAI AI Text Classifier Performance on AI Text.
The bar shows \textbf{26\% True Positive} (Identified) and \textbf{74\% False Negative} (Missed).}
\end{minipage}%
\hfill
\begin{minipage}[t]{0.56\textwidth}
\vspace{0pt}
\begin{tcolorbox}[
  enhanced, breakable,
  title={\textsc{Question} \textnormal{\textit{(Chart Reading, L1 --- AI in the Classroom)}}},
  fonttitle=\small\bfseries\sffamily, coltitle=white,
  colbacktitle=gray!65, colback=gray!4, colframe=gray!40,
  boxrule=0.4pt, arc=2pt, left=5pt, right=5pt, top=2pt, bottom=2pt
]
\footnotesize
According to Figure~14 (`OpenAI AI Text Classifier Performance on AI Text (\%)'),
what percentage of AI-generated text resulted in a
\textbf{False Negative (Missed)}?
\end{tcolorbox}
\smallskip
\begin{tcolorbox}[
  enhanced, breakable,
  title={\textsc{Ground Truth}},
  fonttitle=\small\bfseries\sffamily, coltitle=white,
  colbacktitle=blue!55, colback=blue!3, colframe=blue!30,
  boxrule=0.4pt, arc=2pt, left=5pt, right=5pt, top=2pt, bottom=2pt
]
\footnotesize
\textbf{74\%} of AI-generated text resulted in a False Negative (Missed).
\end{tcolorbox}
\end{minipage}

\medskip
\noindent\textbf{Gemini-3.1-Pro} \scorehi{10/10}:
\begin{tcolorbox}[enhanced,breakable,colback=lbgreen!15,colframe=lbgreen!60,boxrule=0.4pt,arc=2pt,left=6pt,right=6pt,top=2pt,bottom=2pt]
\footnotesize
According to Figure~14, \cor{74\%} of AI-generated text resulted in a False Negative (Missed).
\end{tcolorbox}

\noindent\textbf{GPT-4o} \scorelo{0/10}:
\begin{tcolorbox}[enhanced,breakable,colback=red!5,colframe=red!30,boxrule=0.4pt,arc=2pt,left=6pt,right=6pt,top=2pt,bottom=2pt]
\footnotesize
According to Figure~14, \wrg{43\%} of AI-generated text resulted in a False Negative (Missed).
\end{tcolorbox}

\noindent\textbf{Claude-Sonnet-4.5} \scorehi{10/10}:
\begin{tcolorbox}[enhanced,breakable,colback=lbgreen!15,colframe=lbgreen!60,boxrule=0.4pt,arc=2pt,left=6pt,right=6pt,top=2pt,bottom=2pt]
\footnotesize
Figure~14 shows that \cor{74\%} of AI-generated text was classified as a False Negative (Missed),
meaning the classifier failed to identify it as AI-written.
\end{tcolorbox}

\noindent\textbf{Qwen3-VL-235B} \scorelo{0/10}:
\begin{tcolorbox}[enhanced,breakable,colback=red!5,colframe=red!30,boxrule=0.4pt,arc=2pt,left=6pt,right=6pt,top=2pt,bottom=2pt]
\footnotesize
Based on Figure~14, \wrg{26\%} of AI-generated text resulted in a False Negative (Missed).
\end{tcolorbox}

\noindent\textbf{Qwen2.5-VL-7B} \scorelo{0/10}:
\begin{tcolorbox}[enhanced,breakable,colback=red!5,colframe=red!30,boxrule=0.4pt,arc=2pt,left=6pt,right=6pt,top=2pt,bottom=2pt]
\footnotesize
According to Figure~14, \wrg{26\%} of AI-generated text resulted in a False Negative (Missed).
\end{tcolorbox}

\noindent\textbf{InternVL3-78B} \scorelo{0/10}:
\begin{tcolorbox}[enhanced,breakable,colback=red!5,colframe=red!30,boxrule=0.4pt,arc=2pt,left=6pt,right=6pt,top=2pt,bottom=2pt]
\footnotesize
According to Figure~14, the percentage of AI-generated text that resulted in a
False Negative (Missed) is \wrg{16\%}.
\end{tcolorbox}

\noindent\textit{The chart clearly labels both segments: 26\% (True Positive, blue) and 74\% (False Negative, gray). Only Gemini and Claude read the correct gray segment. GPT-4o hallucinates 43\%---a value absent from the chart. Both Qwen models confuse the two segments, reading the True Positive bar (26\%) as the answer. InternVL3 hallucinates 16\%.}

\clearpage
\subsection*{Example 2 — Cross-Modal L2 \hfill \textit{Failure: Visual Hallucination}}
\noindent\rule{\linewidth}{0.4pt}
\begin{minipage}[t]{0.40\textwidth}
\vspace{0pt}
\centering
\includegraphics[width=\linewidth]{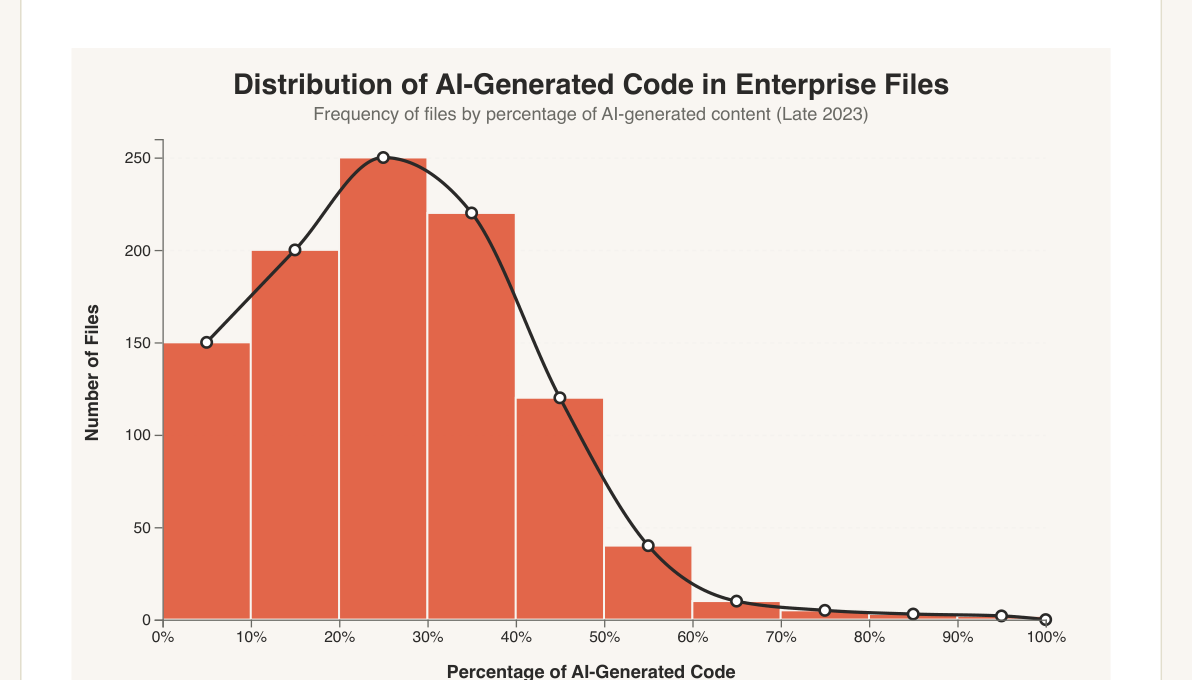}\\[3pt]
\footnotesize\textit{Figure~1 from the document (p.5) --- Distribution of AI-Generated Code in Enterprise Files.
The 40--50\% bin has \textbf{120 files}; the 50--60\% bin has \textbf{40 files}.}
\end{minipage}%
\hfill
\begin{minipage}[t]{0.56\textwidth}
\vspace{0pt}
\begin{tcolorbox}[
  enhanced, breakable,
  title={\textsc{Question} \textnormal{\textit{(Cross-Modal, L2 --- AI-Native Software Dev.)}}},
  fonttitle=\small\bfseries\sffamily, coltitle=white,
  colbacktitle=gray!65, colback=gray!4, colframe=gray!40,
  boxrule=0.4pt, arc=2pt, left=5pt, right=5pt, top=2pt, bottom=2pt
]
\footnotesize
The text describes the distribution dropping `sharply after the 40--50\% mark.'
How does Figure~1 quantify this as the \textbf{exact file frequency drop}
between the 40--50\% and 50--60\% bins?
\end{tcolorbox}
\smallskip
\begin{tcolorbox}[
  enhanced, breakable,
  title={\textsc{Ground Truth}},
  fonttitle=\small\bfseries\sffamily, coltitle=white,
  colbacktitle=blue!55, colback=blue!3, colframe=blue!30,
  boxrule=0.4pt, arc=2pt, left=5pt, right=5pt, top=2pt, bottom=2pt
]
\footnotesize
40--50\% bin: \textbf{120 files} $\to$ 50--60\% bin: \textbf{40 files}.
Drop of \textbf{80 files} (66.7\% reduction).
\end{tcolorbox}
\end{minipage}

\medskip
\noindent\textbf{Gemini-3.1-Pro} \scorelo{0/10}:
\begin{tcolorbox}[enhanced,breakable,colback=red!5,colframe=red!30,boxrule=0.4pt,arc=2pt,left=6pt,right=6pt,top=2pt,bottom=2pt]
\footnotesize
In Figure~1, the 40--50\% bin shows a frequency of \wrg{200 files}.
The subsequent 50--60\% bin drops to a frequency of \wrg{50 files}.
This represents an exact numerical drop of \wrg{150 files} between the two bins.
\end{tcolorbox}

\noindent\textbf{GPT-4o} \scorelo{0/10}:
\begin{tcolorbox}[enhanced,breakable,colback=red!5,colframe=red!30,boxrule=0.4pt,arc=2pt,left=6pt,right=6pt,top=2pt,bottom=2pt]
\footnotesize
Figure~1 indicates that the file frequency drops from approximately \wrg{17\%} in the
40--50\% bin to approximately \wrg{8\%} in the 50--60\% bin, representing a numerical
decline of about \wrg{9 percentage points}.
\end{tcolorbox}

\noindent\textbf{Claude-Sonnet-4.5} \scorelo{1/10}:
\begin{tcolorbox}[enhanced,breakable,colback=red!5,colframe=red!30,boxrule=0.4pt,arc=2pt,left=6pt,right=6pt,top=2pt,bottom=2pt]
\footnotesize
\wrg{Not found in the document.} While Figure~1 shows the distribution and the text
describes a sharp drop after the 40--50\% mark, the specific numerical values for
file frequency in the two bins are not provided in the visible portions of the document.
\end{tcolorbox}

\noindent\textbf{Qwen3-VL-235B} \scorelo{2/10}:
\begin{tcolorbox}[enhanced,breakable,colback=red!5,colframe=red!30,boxrule=0.4pt,arc=2pt,left=6pt,right=6pt,top=2pt,bottom=2pt]
\footnotesize
Figure~1 shows the frequency of files for the 40--50\% bin is approximately
\wrg{175}. For the next bin, 50--60\%, the frequency drops sharply to approximately
\wrg{50}. This represents a numerical drop of about \wrg{125 files}.
\end{tcolorbox}

\noindent\textbf{Qwen2.5-VL-7B} \scorelo{0/10}:
\begin{tcolorbox}[enhanced,breakable,colback=red!5,colframe=red!30,boxrule=0.4pt,arc=2pt,left=6pt,right=6pt,top=2pt,bottom=2pt]
\footnotesize
The figure shows that the number of files drops from approximately \wrg{250} in the
40--50\% bin to around \wrg{150} in the 50--60\% bin.
\end{tcolorbox}

\noindent\textbf{InternVL3-78B} \scorelo{0/10}:
\begin{tcolorbox}[enhanced,breakable,colback=red!5,colframe=red!30,boxrule=0.4pt,arc=2pt,left=6pt,right=6pt,top=2pt,bottom=2pt]
\footnotesize
\wrg{Not found in the document.}
\end{tcolorbox}

\noindent\textit{Every model hallucinates a different plausible-looking but wrong pair of bar heights. GPT-4o conflates file counts with axis percentages. Claude and InternVL3 decline to answer despite the chart being visible on p.5. No model reads the correct 120$\to$40 counts.}

\clearpage
\subsection*{Example 3 — Cross-Modal L3 \hfill \textit{Failure: Figure Not Found}}
\noindent\rule{\linewidth}{0.4pt}
\begin{minipage}[t]{0.40\textwidth}
\vspace{0pt}
\centering
\includegraphics[width=\linewidth]{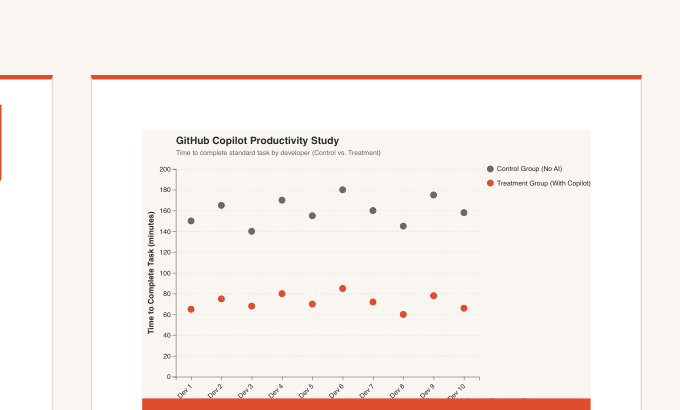}\\[3pt]
\footnotesize\textit{Figure~3 from the document (p.8) --- GitHub Copilot Productivity Study.
\textbf{Dev~6} (control) reaches \textbf{180 min}; Dev~6 (treatment) is at \textbf{85 min}.}
\end{minipage}%
\hfill
\begin{minipage}[t]{0.56\textwidth}
\vspace{0pt}
\begin{tcolorbox}[
  enhanced, breakable,
  title={\textsc{Question} \textnormal{\textit{(Cross-Modal, L3 --- AI-Native Software Dev.)}}},
  fonttitle=\small\bfseries\sffamily, coltitle=white,
  colbacktitle=gray!65, colback=gray!4, colframe=gray!40,
  boxrule=0.4pt, arc=2pt, left=5pt, right=5pt, top=2pt, bottom=2pt
]
\footnotesize
The text states the longest task time in the GitHub Copilot study was 180 min
by a developer in the control group. Using Figure~3, which \textbf{specific
developer} recorded this time, and what was their time in the
\textbf{Treatment Group}?
\end{tcolorbox}
\smallskip
\begin{tcolorbox}[
  enhanced, breakable,
  title={\textsc{Ground Truth}},
  fonttitle=\small\bfseries\sffamily, coltitle=white,
  colbacktitle=blue!55, colback=blue!3, colframe=blue!30,
  boxrule=0.4pt, arc=2pt, left=5pt, right=5pt, top=2pt, bottom=2pt
]
\footnotesize
\textbf{Dev~6} (Control: 180 min $\to$ Treatment: \textbf{85 min}).
\end{tcolorbox}
\end{minipage}

\medskip
\noindent\textbf{Gemini-3.1-Pro} \scorelo{0/10}:
\begin{tcolorbox}[enhanced,breakable,colback=red!5,colframe=red!30,boxrule=0.4pt,arc=2pt,left=6pt,right=6pt,top=2pt,bottom=2pt]
\footnotesize
\wrg{Not found in the document.}
\end{tcolorbox}

\noindent\textbf{GPT-4o} \scorelo{0/10}:
\begin{tcolorbox}[enhanced,breakable,colback=red!5,colframe=red!30,boxrule=0.4pt,arc=2pt,left=6pt,right=6pt,top=2pt,bottom=2pt]
\footnotesize
\wrg{The document does not specifically identify which developer recorded the
180-minute maximum time nor their completion time in the Treatment Group.}
\end{tcolorbox}

\noindent\textbf{Claude-Sonnet-4.5} \scorelo{0/10}:
\begin{tcolorbox}[enhanced,breakable,colback=red!5,colframe=red!30,boxrule=0.4pt,arc=2pt,left=6pt,right=6pt,top=2pt,bottom=2pt]
\footnotesize
\wrg{Not found in the document.} While the document mentions the 180-minute maximum
completion time in the control group, it does not identify which specific developer
recorded this time or provide their individual completion time in the treatment group.
\end{tcolorbox}

\noindent\textbf{Qwen3-VL-235B} \scorelo{0/10}:
\begin{tcolorbox}[enhanced,breakable,colback=red!5,colframe=red!30,boxrule=0.4pt,arc=2pt,left=6pt,right=6pt,top=2pt,bottom=2pt]
\footnotesize
\wrg{Not found in the document.}
\end{tcolorbox}

\noindent\textbf{Qwen2.5-VL-7B} \scorelo{0/10}:
\begin{tcolorbox}[enhanced,breakable,colback=red!5,colframe=red!30,boxrule=0.4pt,arc=2pt,left=6pt,right=6pt,top=2pt,bottom=2pt]
\footnotesize
\wrg{Not found in the document.}
\end{tcolorbox}

\noindent\textbf{InternVL3-78B} \scorelo{0/10}:
\begin{tcolorbox}[enhanced,breakable,colback=red!5,colframe=red!30,boxrule=0.4pt,arc=2pt,left=6pt,right=6pt,top=2pt,bottom=2pt]
\footnotesize
\wrg{Not found in the document.}
\end{tcolorbox}

\noindent\textit{Figure~3 is clearly visible on p.8 with per-developer labels on the x-axis (Dev~1 through Dev~10). The highest gray dot at 180 min sits at Dev~6. All six evaluated models fail to localise and resolve the individual data point label, returning a uniform ``not found'' despite the chart being visually present.}

\clearpage
\subsection*{Example 4 — Complex Multi-Hop L5 \hfill \textit{Failure: Incomplete Retrieval}}
\noindent\rule{\linewidth}{0.4pt}
\begin{minipage}[t]{0.40\textwidth}
\vspace{0pt}
\centering
\includegraphics[width=\linewidth]{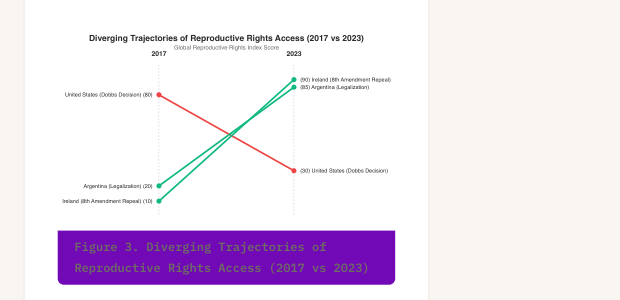}\\[4pt]
\includegraphics[width=\linewidth]{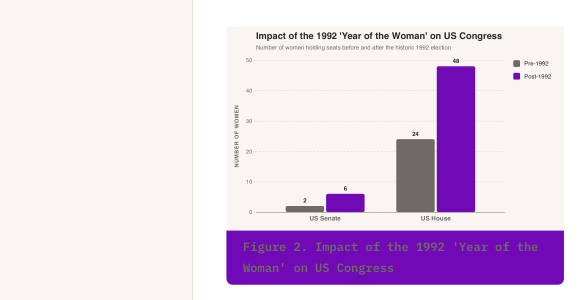}\\[3pt]
\footnotesize\textit{Two of three evidence figures from p.5 of the gender equality document:
Figure~3 (Reproductive Rights Trajectories, top) and the Congress representation figure (bottom).
A third chart on the same page shows child marriage rates 2013--2023.}
\end{minipage}%
\hfill
\begin{minipage}[t]{0.56\textwidth}
\vspace{0pt}
\begin{tcolorbox}[
  enhanced, breakable,
  title={\textsc{Question} \textnormal{\textit{(Complex Multi-Hop, L5 --- Gender Equality)}}},
  fonttitle=\small\bfseries\sffamily, coltitle=white,
  colbacktitle=gray!65, colback=gray!4, colframe=gray!40,
  boxrule=0.4pt, arc=2pt, left=5pt, right=5pt, top=2pt, bottom=2pt
]
\footnotesize
How did women's representation in US Congress change after the 1992 `Year of the
Woman'? What was the exact decline in global child marriage rates 2013--2023?
How does the US reproductive rights 50-point drop compare to changes in
\textbf{Ireland} and \textbf{Argentina} over 2017--2023?
\end{tcolorbox}
\smallskip
\begin{tcolorbox}[
  enhanced, breakable,
  title={\textsc{Ground Truth}},
  fonttitle=\small\bfseries\sffamily, coltitle=white,
  colbacktitle=blue!55, colback=blue!3, colframe=blue!30,
  boxrule=0.4pt, arc=2pt, left=5pt, right=5pt, top=2pt, bottom=2pt
]
\footnotesize
House: \textbf{24$\to$48} (doubled). Senate: \textbf{2$\to$6} (tripled).\\
Child marriage: \textbf{25$\to$20} (drop of 5).\\
US: \textbf{80$\to$30}. Ireland: \textbf{10$\to$90}. Argentina: \textbf{20$\to$85}.
\end{tcolorbox}
\end{minipage}

\medskip
\noindent\textbf{Gemini-3.1-Pro} \scoremid{4/10}:
\begin{tcolorbox}[enhanced,breakable,colback=red!5,colframe=red!30,boxrule=0.4pt,arc=2pt,left=6pt,right=6pt,top=2pt,bottom=2pt]
\footnotesize
Following the 1992 election, voters elected 4 new female senators (bringing the
Senate total to \cor{6}) and \cor{24} new women to the House of Representatives.
Between 2013 and 2023, global child marriage rates declined from
\cor{25\%} to \wrg{19\%}.
During the 2017--2023 timeframe, the 50-point drop in the US Reproductive Rights
Index (from \wrg{75} to \wrg{25}) contrasted with a 60-point increase in Ireland
(from \wrg{20} to \wrg{80}) and a 45-point increase in Argentina
(from \wrg{30} to \wrg{75}).
\end{tcolorbox}

\noindent\textbf{GPT-4o} \scorelo{2/10}:
\begin{tcolorbox}[enhanced,breakable,colback=red!5,colframe=red!30,boxrule=0.4pt,arc=2pt,left=6pt,right=6pt,top=2pt,bottom=2pt]
\footnotesize
Following the 1992 `Year of the Woman' election, the number of women in the
U.S. Congress \cor{doubled}. The global child marriage rate
\wrg{declined by 15\%} between 2013 and 2023.
While the U.S. reproductive rights index dropped by \cor{50 points} from 2017 to 2023,
during the same timeframe, Ireland's index rose by \wrg{25 points} and
Argentina's by \wrg{30 points}.
\end{tcolorbox}

\noindent\textbf{Claude-Sonnet-4.5} \scorelo{0/10}:
\begin{tcolorbox}[enhanced,breakable,colback=red!5,colframe=red!30,boxrule=0.4pt,arc=2pt,left=6pt,right=6pt,top=2pt,bottom=2pt]
\footnotesize
\wrg{Not found in the document.}
\end{tcolorbox}

\noindent\textbf{Qwen3-VL-235B} \scorelo{3/10}:
\begin{tcolorbox}[enhanced,breakable,colback=red!5,colframe=red!30,boxrule=0.4pt,arc=2pt,left=6pt,right=6pt,top=2pt,bottom=2pt]
\footnotesize
Following the 1992 `Year of the Woman', the number of women in the US Congress
increased from \wrg{4} to \wrg{6}. Global child marriage rates declined from
\cor{25\%} in 2013 to \cor{20\%} in 2023, a \cor{5 percentage point} drop.
The US reproductive rights index dropped \cor{50 points} from 2017 to 2023, while
Ireland's index rose from \wrg{70} to \cor{90} and Argentina's rose from
\wrg{60} to \cor{85} during the same period.
\end{tcolorbox}

\noindent\textbf{Qwen2.5-VL-7B} \scorelo{2/10}:
\begin{tcolorbox}[enhanced,breakable,colback=red!5,colframe=red!30,boxrule=0.4pt,arc=2pt,left=6pt,right=6pt,top=2pt,bottom=2pt]
\footnotesize
The number of women in the US Congress increased from \wrg{4} to \wrg{44} after the
1992 `Year of the Woman' election. The global child marriage rate declined by
approximately \wrg{25 percentage points} between 2013 and 2023.
The US reproductive rights index dropped by \cor{50 points}, while Ireland saw a
\wrg{10-point increase} and Argentina experienced a \wrg{10-point decrease}
during the same period.
\end{tcolorbox}

\noindent\textbf{InternVL3-78B} \scoremid{4/10}:
\begin{tcolorbox}[enhanced,breakable,colback=red!5,colframe=red!30,boxrule=0.4pt,arc=2pt,left=6pt,right=6pt,top=2pt,bottom=2pt]
\footnotesize
Following the 1992 `Year of the Woman' election, the number of women elected to
the U.S. House of Representatives increased by \cor{24}, marking a then-record high.
The global child marriage rate declined by \wrg{12 percentage points} between 2013
and 2023. The U.S. reproductive rights index experienced a \cor{50-point drop},
contrasting with Ireland and Argentina, which saw \wrg{significant improvements}
in their index scores during the 2017--2023 period.
\end{tcolorbox}

\noindent\textit{The question requires six independent statistics from three charts on the same page. Gemini retrieves the Senate total and House count but uses wrong base scores for all three countries. Qwen3 correctly reads the child marriage drop and US delta but has wrong starting values for Ireland and Argentina. GPT-4o correctly identifies the Congress doubling and US 50-point drop but reports wrong deltas for the other countries. InternVL3 correctly retrieves the House count and US drop but vaguely describes Ireland/Argentina without numbers. Claude fails to retrieve any evidence. The Ireland (10$\to$90) and Argentina (20$\to$85) base scores are consistently missed.}

\end{document}